\documentclass[sigconf]{acmart}

\usepackage{amsmath}
\usepackage{subcaption}
\usepackage{booktabs}
\usepackage{flushend}
\setcitestyle{numbers}

\copyrightyear{2018}
\acmYear{2018}
\setcopyright{acmcopyright}
\acmConference[ASPLOS '18]{2018 Architectural Support for Programming Languages and Operating Systems}{March 24--28, 2018}{Williamsburg, VA, USA}
\acmBooktitle{ASPLOS '18: 2018 Architectural Support for Programming Languages and Operating Systems, March 24--28, 2018, Williamsburg, VA, USA}
\acmPrice{15.00}
\acmDOI{10.1145/3173162.3173212}
\acmISBN{978-1-4503-4911-6/18/03}
\begin{document}
\title{VIBNN: Hardware Acceleration of Bayesian Neural Networks}

\author{Ruizhe Cai}
\authornote{Ruizhe Cai and Ao Ren contributed equally to this work.}

\author{Ao Ren}
\authornotemark[1]

\author{Ning Liu}

\author{Caiwen Ding}

\affiliation{
            \institution{Department of Electrical Engineering and Computer Science, Syracuse University}
            \city{Syracuse}
            \state{New York}
            \postcode{13244}
            }
\email{{rcai100,aren,nliu03,cading}@syr.edu}

\author{Luhao Wang}

\author{Xuehai Qian}

\author{Massoud Pedram}

\affiliation{
            \institution{Department of Electrical Engineering, University of Southern California}
            \city{Los Angeles}
            \state{California}
            \postcode{90089}
            }
\email{{luhaowan,xuehai.qian,pedram}@usc.edu}

\author{Yanzhi Wang}
\affiliation{
            \institution{Department of Electrical Engineering and Computer Science, Syracuse University}
            \city{Syracuse}
            \state{New York}
            \postcode{13244}
            }
\email{ywang393@syr.edu}

\renewcommand{\shortauthors}{B. Trovato et al.}

\newcommand\numberthis{\addtocounter{equation}{1}\tag{\theequation}}
\renewcommand{\shortauthors}{R. Cai and A. Ren et al.}

\begin{abstract}
Bayesian Neural Networks (BNNs) have been proposed to address the problem of model uncertainty in training and inference. By introducing weights associated with conditioned probability distributions, BNNs are capable of resolving the overfitting issue 
commonly seen in conventional neural networks and allow for small-data training, through the variational inference process. Frequent usage of Gaussian random variables in this process requires a properly optimized Gaussian Random Number Generator (GRNG). The high hardware cost of conventional GRNG makes the hardware implementation of BNNs challenging. 

In this paper, we propose VIBNN, 
an FPGA-based hardware accelerator design for variational inference on BNNs. 
We explore the design space for massive amount of  Gaussian variable sampling tasks in BNNs.
Specifically, we introduce two high performance Gaussian (pseudo) random number generators: 
{\em 1)} the RAM-based Linear Feedback Gaussian Random Number Generator (RLF-GRNG), 
which is inspired by the properties of binomial distribution and linear feedback logics; and 
{\em 2)} the Bayesian Neural Network-oriented Wallace Gaussian Random Number Generator. 
To achieve high scalability and efficient memory access,
we propose a deep pipelined accelerator architecture 
with fast execution and good hardware utilization.  
Experimental results demonstrate that the proposed VIBNN implementations on an FPGA can achieve throughput of 321,543.4 Images/s and energy efficiency upto 52,694.8 Images/J while maintaining similar accuracy as its software counterpart.
\end{abstract}
%
%


\maketitle

\section{Introduction}
As a key branch of machine learning and artificial intelligence techniques, Artificial Neural Networks (ANNs) have been introduced to create machines that can learn and inference \cite{Goodfellow-et-al-2016}. Many different types and models of ANNs have been developed for a variety of applications and higher performance, including \emph{Convolutional Neural Networks} (CNNs), \emph{Multi-Layer Perceptron Networks} (MLPs), \emph{Recurrent Neural Networks} (RNNs), etc. \cite{schmidhuber2015deep}. With the development and broad applications of deep learning algorithms, neural networks have recently achieved tremendous success in various fields, such as image classification, object recognition, natural language processing, autonomous driving, cancer detection, etc. \cite{simonyan2014very,agrawal2014analyzing,sutskever2014sequence,cirecsan2013mitosis}. 

With the success of deep learning, a rising amount of recent works studied the highly parallel computing paradigm and the hardware implementations of neural networks\cite{nurvitadhi2017can,suda2016throughput,akopyan2015truenorth,ren2017sc,jouppi2017datacenter,chen2014dadiannao,chen2017eyeriss,han2016eie}. These hardware approaches typically accelerate the inference process of neural networks 
and have shown promising performances in terms of speed, energy efficiency, and accuracy, 
making this approach ideal for embedded and IoT systems. 
Despite the significant progress of neural network acceleration , 
it is well known that conventional neural networks are prone to the \emph{over-fitting issue} 
--- situations where the model fail to generalize well from the training data to the test data \cite{gal2015bayesian}. 
The fundamental reason is that traditional neural network models fail to provide estimates with uncertainty information \cite{blundell2015weight}. This missing characteristic is crucial to avoid making over-confident decisions, 
especially for many supervised learning applications with missing or noisy training data. 

To solve this issue, the \emph{ensemble model} has been introduced \cite{dietterich2000ensemble,ghahramani2001propagation} 
to combine the results from multiple neural network models, so that 
the degraded generalization performance can be avoided. 
As a key example, \textit{Bayesian Neural Network} (BNNs) are capable of 
forming ensemble models while maintaining limited memory space overhead \cite{liu1999ensemble,zhang2012ensemble}. 
In addition, unlike conventional neural networks that rely on huge amount of data for training, 
BNNs can easily learn from small datasets, 
with the ability to offer uncertainty estimates and the robustness to mitigate over-fitting issues\cite{gal2015bayesian}. 
Moreover, the overall accuracy can be improved as well.   

Specifically, BNNs apply Bayesian inference to provide the principled uncertainty estimates. 
In contrast to traditional neural networks whose weights are fixed values, 
each weight in BNNs is a {\em random number} following a 
\emph{posterior probability distribution}, 
which is conditioned on a \emph{prior probability} and its \emph{observed data}. 
Unfortunately, the exact Bayesian inference in general is an intractable problem and obtaining 
closed-form solutions requires either the assumption of special families of models\cite{braun2010variational}
or the availability of probability distributions \cite{ghahramani2001propagation}. 
Therefore, an {\em approximation method of Bayesian inference} is generally used to ensure 
low computational complexity and high degree of generality \cite{blei2010nested,teh2010hierarchical}. 
Among various approximation techniques, \emph{variational approximation}, or \emph{variational inference}, 
tends to be faster and easier to implement. Besides, the variational inference method offers better scalability with large models, especially for large-scale neural networks in deep learning applications\cite{houthooft2016vime},
compared with other commonly adopted Bayesian inference methods such as Markov Chain Monte Carlo (MCMC) \cite{andrieu2003introduction}. 
In addition to faster computation, 
the variational inference method can efficiently represent weights (probability distributions) 
with limited amount of parameters. The Bayes-by-Backprop algorithm proposed by Bluendell 
{\em et al.}\cite{blundell2015weight}, for instance, only doubles the parameters compared 
to ANNs while achieving an infinitely large ensemble of models.      

Our focused BNNs in this work belong to the category of {\em feed-forward neural networks (FNNs)} 
that have achieved great successes in many important fields, 
such as the HIGGS challenge, the Merck Molecular Activity challenge, and the Tox21 Data challenge \cite{klambauer2017self}.
Despite the tremendous attentions on CNNs and RNNs,
the accelerators for FNN models are imperative as well, 
which is noted in the recent Google paper \cite{jouppi2017datacenter} and the very recently invented SeLU technique \cite{klambauer2017self}. 
With the recent shift in various fields towards the deployment of BNNs \cite{fortunato2017bayesian,blundell2015weight,ticknor2013bayesian}, 
hardware acceleration for BNNs becomes critical and has not been well considered in prior works. 
However, hardware realizations of BNNs pose a fundamental challenge compared to 
the traditional ANNs: the frequent operations on {\em probability distributions} requires 
additional logic circuits designed and optimized for (Gaussian) random number generation. 

In this paper, we propose {\em VIBNN}, 
an FPGA-based hardware accelerator design for variational inference on BNNs. 
In VIBNN, posterior distributions of network weights are approximated as 
Gaussian distributions associated with variational parameters (trainable mean values and variances). 
As a common practice, network training is performed 
on CPU/GPU clusters before the parameters for weights distributions. 
We explore the design space for massive amount of  Gaussian variable sampling tasks in BNNs.
Specifically, we introduce two high performance Gaussian (psuedo) random number generators: 
{\em 1)} the RAM-based Linear Feedback Gaussian Random Number Generator (RLF-GRNG), 
which is inspired by the properties of binomial distribution and linear feedback logics; and 
{\em 2)} the Bayesian Neural Network-oriented Wallace Gaussian Random Number Generator. 
To achieve high scalability and efficient memory access,
we propose a deep pipelined accelerator architecture 
with fast execution and good hardware utilization.  
It is important to note that
BNN is a mathematical model, instead of a specific type of neural network structure.
Therefore, the design principles of VIBNN 
are orthogonal to the optimization techniques on convolutional layers in previous works \cite{mathieu2013fast,denil2013predicting,jaderberg2014speeding},
and can be applied to CNNs and RNNs as well.

Experimental results suggest that the proposed VIBNN can achieve similar accuracy as its software counterpart (BNNs) with very high energy efficiency of $52694.9$ images/J thanks to the proposed GRNG structure.   
\section{BNNs using Variational Inference}
\subsection{Bayesian Model, Variational Inference, and Gaussian Approximation}
\label{sec:vib}
For a general Bayesian model, the \emph{latent variables} are $\mathbf{w}$ and \emph{observed data points} are $\mathcal{D}$. From the Bayes rule, the \emph{posterior probability} can be calculated as:
\begin{equation}
\label{pos1}
P(\mathbf{w}|\mathcal{D}) = \frac{P(\mathcal{D}|\mathbf{w}) P(\mathbf{w})}{P(\mathcal{D})}
\end{equation}
where $P(\mathbf{w})$ is called the \emph{prior probability} that indicates the probability of latent variables $\mathbf{w}$ before any data observations. $P(\mathcal{D}|\mathbf{w})$ is called the \emph{likelihood}, which is the probability of the data $\mathcal{D}$ based on the latent variable observations $\mathbf{w}$. The denominator $P(\mathcal{D})$ is calculated as the integral of sum over all possible latent variables, i.e., $P(\mathcal{D}) = \int P(\mathcal{D}|\mathbf{\mathbf{w}}) P(\mathbf{w}) d\mathbf{w}$. 

For most of applications of interests, this integral process is intractable, therefore effective approaches are needed to {\em approximately} estimate/evaluate the posterior probability. \emph{Variational inference}\cite{jordan1999introduction,wainwright2008graphical} is a machine learning method for approximating the \emph{posterior probability} densities in Bayesian inference models, with a higher convergence rate (compared to MCMC method) and scalability to large problems. As shown in \cite{ghahramani2001propagation}, the variational inference method posits a family of probability distributions $q(\mathbf{w};\theta)$ with \emph{variation parameters} $\theta$ to approximate the posterior distribution $p(\mathbf{w}|\mathcal{D})$. 

For simplicity, one can assume that {the variational posterior distribution is a Gaussian distribution}, then the variational parameters (vector) $\theta$ can be specified as $\theta = (\mu, \rho)$, where $\mu$ is the vector of mean values and $\rho$ is used to further produce the non-negative vector of standard deviations, i.e., $\sigma = \ln (1+exp(\rho))$. Therefore, a sample of $\mathbf{w}$ can be obtained by shifting and scaling \emph{unit Gaussian variables}:
\begin{equation}
\label{eq:tran}
\mathbf{w} = \mu+\epsilon \circ \ln (1+exp(\rho))
\end{equation}
where $\epsilon \sim \mathcal{N}(0, \mathbf{I})$ is a vector of independent unit Gaussian variables, and $\circ$ denotes the element-wise multiplication operation. Therefore, \emph{generating unit Gaussian variables is the key step in generating samples of $\mathbf{w}$.}

\subsection{Gaussian Variational Inference with a BNN}
\begin{figure}[t]
\centering
\includegraphics[width=0.3\textwidth]{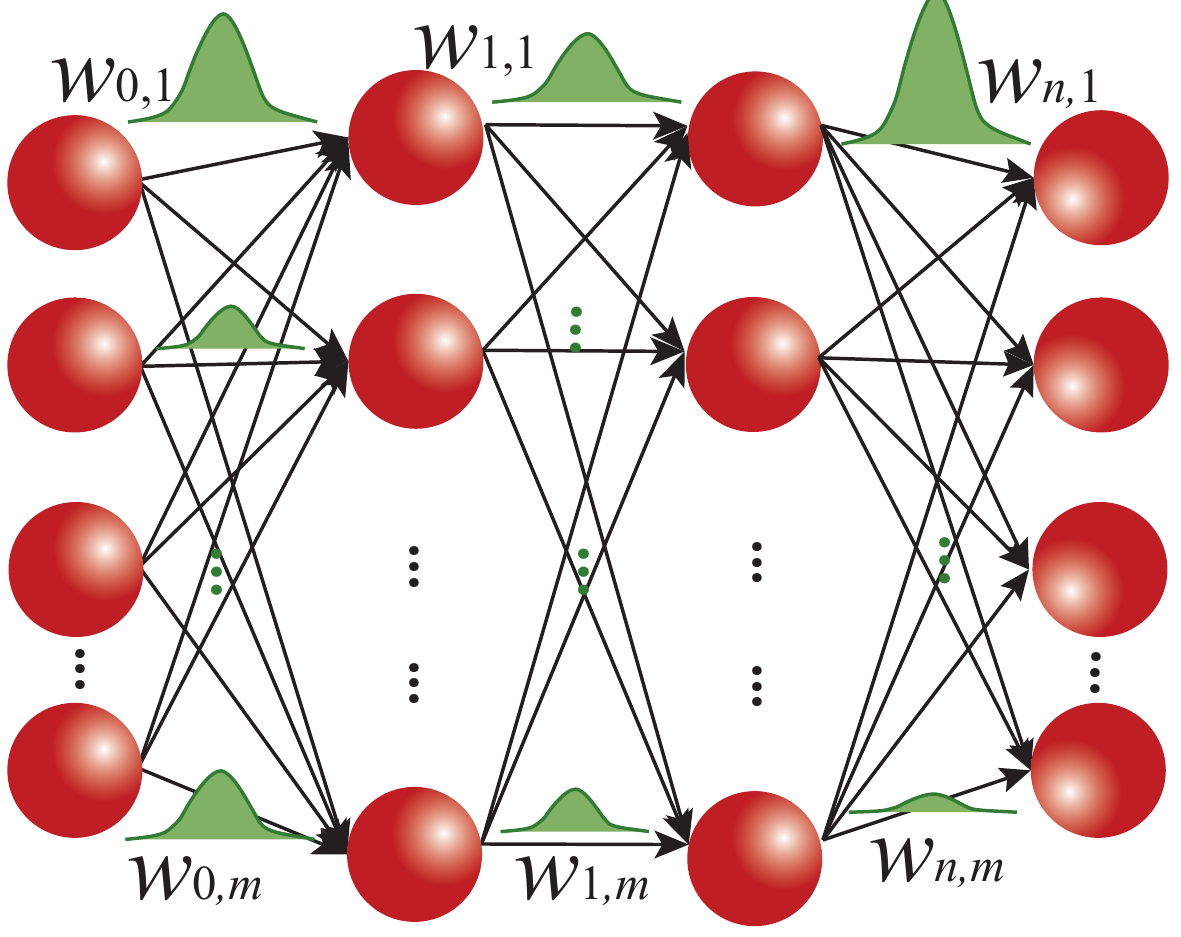}
\caption{Illustration of model structure of BNNs.}
\label{fig:bnn}
\end{figure}

\emph{Bayesian Neural Networks} (BNNs) are first suggested in the 1990s and studied extensively since then \cite{mackay1992bayesian, neal2012bayesian}. Compared to conventional neural networks with fixed value weights representing deterministic models, BNNs offer ensembles of models by 
\emph{incorporating probability distributions over models' weights}, as shown in Figure \ref{fig:bnn}. As a result, BNNs achieve higher overall accuracy and are more robust to over-fitting. In addition, unlike conventional neural networks that require large data sets to achieve good accuracy, 
BNNs can easily learn from small data with much improved convergence rate. 

This paper considers Gaussian variational inference for BNNs. 
The goal is to perform inference using a BNN with learned posterior distribution $p(\mathbf{w}|\mathcal{D})$ of network weights.
To fully utilize the posterior distribution, the network output should be derived by \emph{averaging over the outputs produced according to the posterior distribution.} In other words, the output of network, $\mathbf{y}$, is calculated as:
\begin{equation}
\mathbf{y}=\mathbb{E}_{p(\mathbf{w}|\mathcal{D})}[g(\mathbf{x}_0, \mathbf{w})]
\end{equation}
where $g(\mathbf{x}_0, \mathbf{w})$ is the network function and $\mathbf{x}_0$ is the network input. With the variational inference method, the network output can be represented as:
\begin{equation}
\mathbf{y}\approx \mathbb{E}_{q(\mathbf{w};\theta)}[g(\mathbf{x}_0, \mathbf{w})]
\end{equation}
Using a Monte Carlo Sampling, the output can be further approximated as:
\begin{equation}
\mathbf{y}\approx  \frac{1}{N}\sum_{i=1}^{N}g(\mathbf{x}_0, \mathbf{w}^i)
\end{equation}
where $N$ is the total sample count and $\mathbf{w}^i$ is the $i$-th Monte Carlo sample drawn from the variational posterior distribution $q(\mathbf{w};\theta)$, which can be obtained using equation (\ref{eq:tran}). Then the output is estimated as:
\begin{equation}
\mathbf{y}\approx  \frac{1}{N}\sum_{i=1}^{N}g(\mathbf{x}_0, \mu+\sigma\circ \epsilon^i)
\end{equation}
where $\epsilon^i$ is the $i$-th Monte Carlo sample of unit Gaussian variables. As shown in Figure \ref{fig:bnn}, each weight requires sampling a unit Gaussian random variable. Then a BNN can perform inference using the same method as a normal ANN. 

Since our target implementation platforms are low-power, embedded systems based on FPGAs,
the network is trained offline, as a widely adopted practice in hardware deep learning research~\cite{farabet2011neuflow,ren2017sc, chen2014dadiannao,akopyan2015truenorth,ovtcharov2015accelerating,wen2016new}, 
using high performance computing platforms such as GPUs.
Afterward, the trained variational parameters (vectors) $\mu$ and $\sigma$ 
are migrated to the memory of the target FPGA platform.

\subsection{Gaussian Random Number Generators (GRNGs)}
As introduced before, weights and biases in the BNNs of interests are drawn from Gaussian distributions $\mathcal{N}(\mu,\sigma^2)$.
Therefore, it is crucial to rapidly generate high-quality Gaussian random numbers. 

The methods of generating \emph{Gaussian Random Numbers} (GRNs) can be classified into four categories: 
{\em 1)} the \emph{cumulative density function} (CDF) inversion method that obtains GRNs simply by inverting the CDFs, such as \cite{beasley1985percentage,muller1958inverse}; 
{\em 2)} the \emph{transformation method} that obtains GRNs through a series of operations on uniform distributions, according to the \emph{Central Limit Theorem} (CLT), such as \cite{teichroew1953distribution, muller1959comparison}; 
{\em 3)} the \emph{rejection method} that generates GRNs based on the transformation method, but with one additional rejection step that conditionally rejects some of the transformed numbers, such as the Ziggurat algorithm \cite{marsaglia1984fast}; 
{\em 4)} the \emph{recursion method} that generates GRNs by combining previously generated GRNs in a linear manner, such as the Wallace method \cite{wallace1996fast}.

For hardware implementations, not all algorithms that are successful on software are appropriate. This is because of the restrictions such as hardware resources, power/energy constraints, and implementation complexity. 
Therefore, the appropriate selection of hardware-based GRNGs 
is typically application-specific and needs to be carefully investigated. 
In this paper, we believe the 
\emph{CLT-based methods} and the {\em Wallace method} 
to be the most appropriate choices for hardware neural network acceleration. 
The main consideration is the lower computation overhead, which 
facilitates the efficient hardware implementation of BNNs.

One of the major advantages of the CLT-based GRNGs is
that uniform random numbers can be easily generated with 
\emph{linear-feedback shift registers} (LFSRs). 
However, the generation quality is affected by various factors such as
the number of stages in LFSRs, the bit-width, etc. \cite{malik2016gaussian}. 
Being another method with low computation overhead,
Wallace algorithm \cite{malik2016gaussian} guarantees the 
correctness by the fact that a linear combination of Gaussian 
random numbers is still a Gaussian random number \cite{wallace1996fast}.
Nevertheless, Wallace algorithm has two main drawbacks: 
First, an initial pool of Gaussian random numbers is needed, thereby adding requirements of memory storage; Second, the generated random numbers have correlations to some degree. 
We propose novel hardware implementation of the CLT-based and Wallace method,
to sufficiently exploit the advantages of these GRNG methods 
but at the same time mitigate the drawbacks.
Moreover, a series of optimizations are performed to further reduce the hardware cost and meanwhile guarantee the high quality of GRNs generated. 

\begin{figure}[b]
\centering
\includegraphics[width=0.4\textwidth]{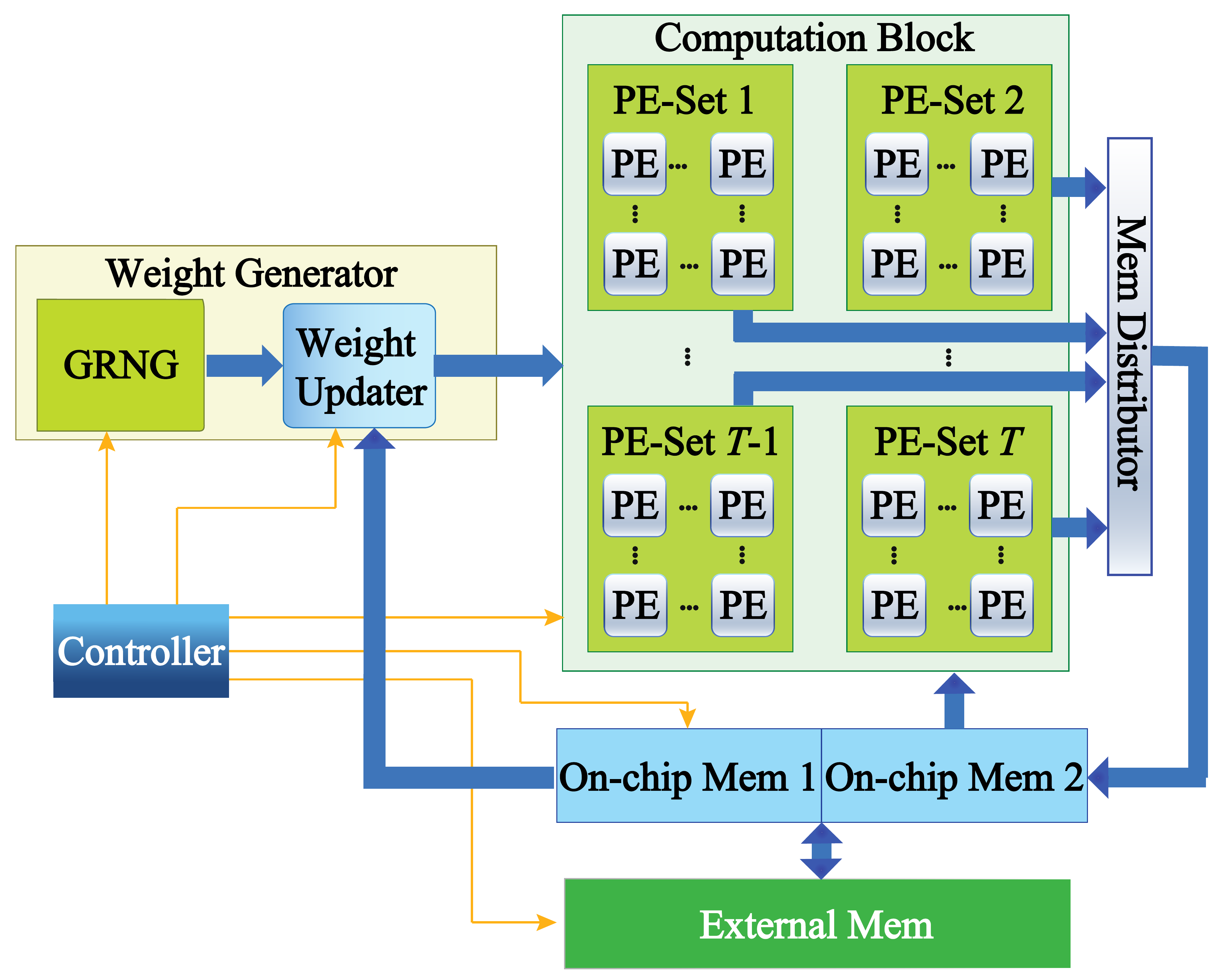}
\caption{Overall architecture of VIBNN}
\label{framework_overview}
\end{figure}

\section{VIBNN Architecture Overview}
\label{sec:grng}
In this section, we present the overview of our VIBNN architecture. As shown in Fig. \ref{framework_overview}, the architecture of VIBNN accelerator
consists of the external memory, on-chip memories, 
a computation block, a weight generator and a global controller. 
The external memory initially stores the input features to the input layer and weight parameters $(\mu, \sigma)$. These inputs and weights are then loaded into on-chip memories 
for online inference. There are two types of on-chip memories: 
one stores the input features, intermediate results and the inference results; 
the other stores the weight parameters. 

As discussed in Section \ref{sec:vib}, the weights that actually participate in the variational inference process are calculated by equation (\ref{eq:tran}). 
Hence, a \emph{weight generator} that comprises a GRNG and a \emph{weight updater} is needed. The GRNG generates the random numbers $\epsilon's$, and the weight updater is responsible for implementing the weight updating equation. The computation block comprises groups of \emph{processing elements (PEs)}, each consisting of a number of PEs. 
To improve memory-access efficiency, the \emph{memory distributor} 
in computation block collects the outputs of PEs and performs necessary operations before writing back to on-chip memories. 
The PEs work in a time-multiplexed manner to perform all computations 
in the whole neural network. 
To correctly implement the overall functionality of VIBNN,
the operations of all components are 
controlled by a \emph{global controller}.
Overall, the VIBNN architecture has three key benefits: 
scalability, portability, and memory-access efficiency. 

\section{BNN-Oriented GRNGs}
\label{grng_design}
In this section, we explore the design of GRNGs 
suitable for BNN hardware implementation.
Specifically, we propose two parallel GRNG designs tailored for BNNs:
{\em 1)} {\em RAM-based Linear Feedback GRNG (RLF-GRNG)}, which is based on
a CLT-based method inspired by the \emph{binomial distribution}.
This method is capable of approximating 
a Gaussian distribution \cite{box1978statistics} when the sample size is large enough. 
Importantly, it can be effectively implemented 
{\em using RAM with compact additional logics} for indexing and computing. 
In addition, the control module can be shared among parallel GRNGs. 
{\em 2)} {\em BNN-Oriented Wallace GRNG}, which is based on Wallace method. 
The proposed design substantially mitigated the two main drawbacks,
i.e., a large initial pool requiring large memory block and 
multi-loop transformations that result in long latency, 
making the Wallace algorithm suitable for hardware BNNs. 

\begin{figure}[b]
	\centering
   \begin{minipage}{1\columnwidth}
    	\centering
    	\includegraphics[width=0.7\textwidth]{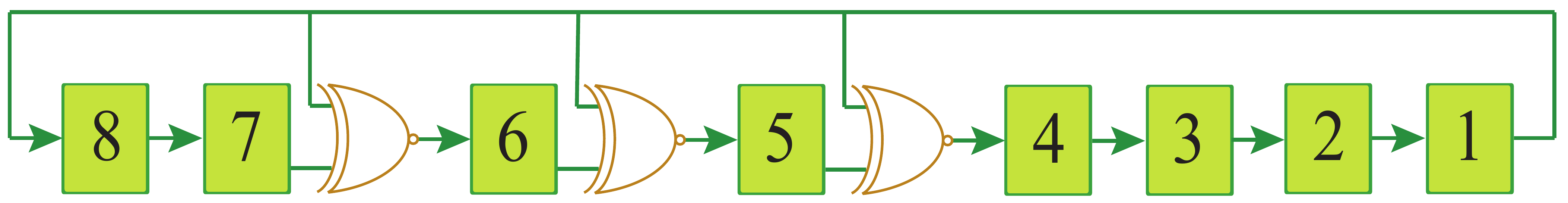}
        \captionof{subfigure}{8-bit LFSR (Register 1 is the head, register 4, 5, 6 are taps)}
   		\centering
    	\includegraphics[width=0.5\textwidth]{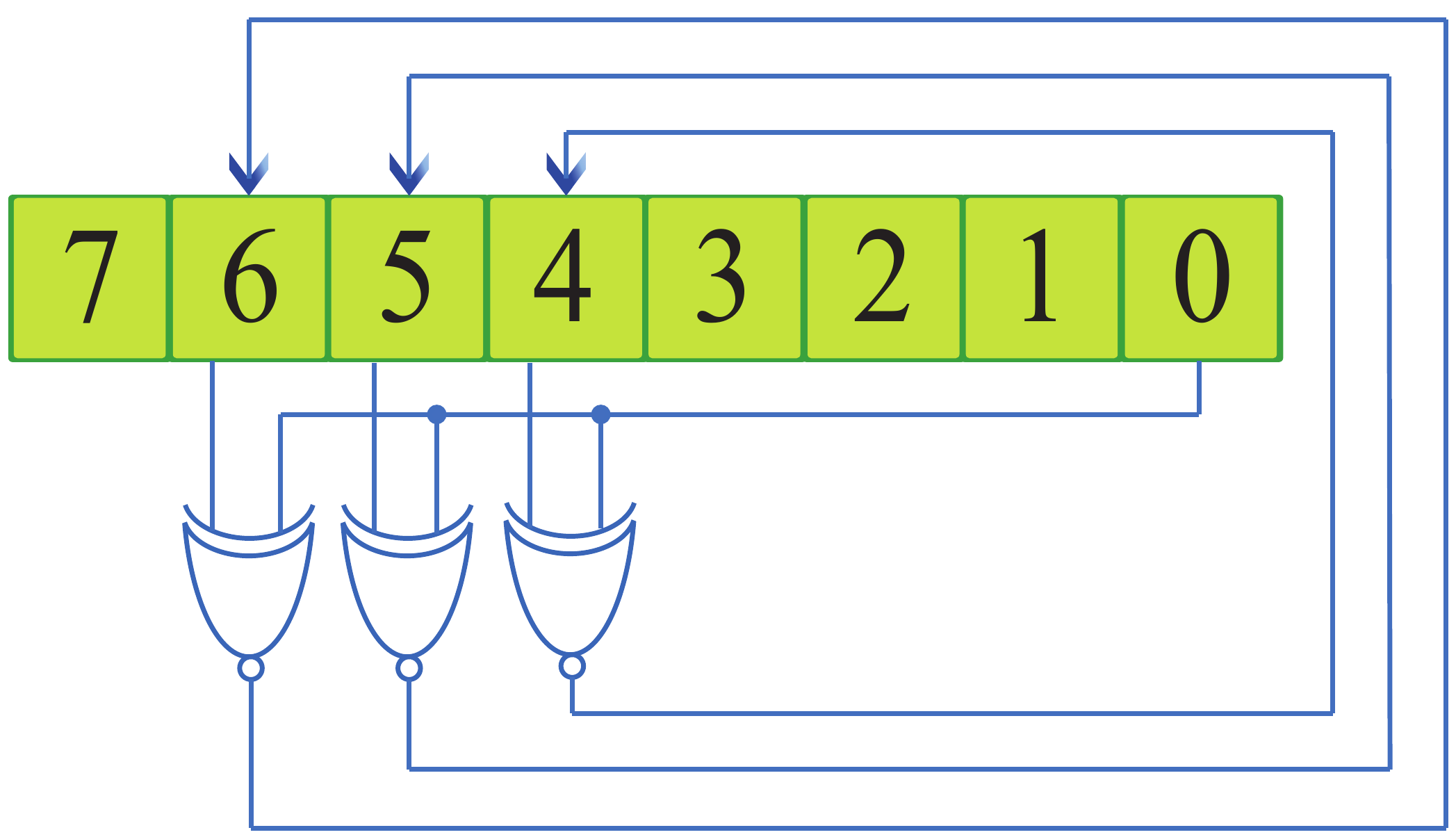}
        \captionof{subfigure}{Equivalent 8-bit RLF Logic}
    \end{minipage}
    \caption{Examples of LFSR and RAM based Linear Feedback. }
	\label{fig:exam}
\end{figure}

\begin{figure}[b]
\centering
\includegraphics[width=0.2\textwidth]{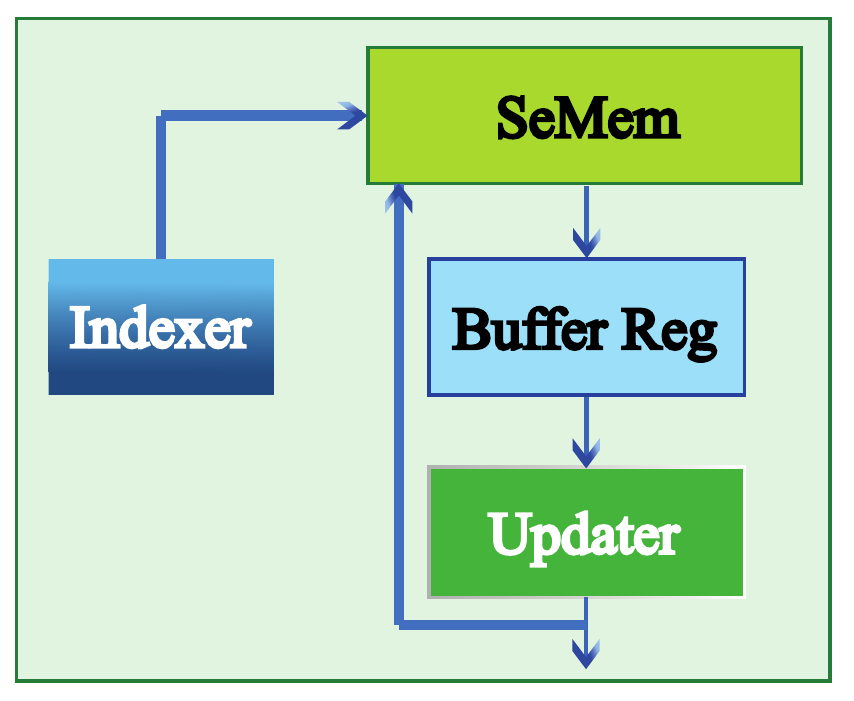}
\caption{Block diagram of the RLF logic}
\label{Fig:rlflogic}
\end{figure}

\subsection{RLF-GRNG Design}

\subsubsection{The Binomial Distribution Approximation Method}
A binomial distribution, denoted as $X \sim{ \mathrm{B}(n,p)}$, is a discrete probability distribution of the number of successes in a sequence of $n$ independent experiments, each with a boolean valued outcome with the probability of success $p$. The probability of getting exact $k$ successes in $n$ trials is given by the following probability mass function:
\begin{equation}
f(k;n,p)=Pr(X=k)=\begin{pmatrix}n\\k\end{pmatrix}p^k{(1-p)}^{n-k}
\end{equation} 
for $k = 0, 1, 2, \dots, n,$ where
$\begin{pmatrix}n\\k\end{pmatrix} = \frac{n!}{k!{(n-k)}!}$.

If $n$ is large enough, $\mathrm{B}(n, p)$ can reasonably approximate a normal distribution: $\mathcal{N}(np, np(1-p))$. More specifically, $n$ is considered to be large enough \cite{walker1985moivre} if:
\begin{equation}
n>9\cdot\frac{1-p}{p}\text{ and }n>9\cdot\frac{p}{1-p}
\label{b2ncon}
\end{equation}

As a simple example, the summation of $n$ individual bits should follow a binomial distribution $\mathrm{B}(n, 0.5)$ if each bit has equal chances of being 0 or 1. According to equation ({\ref{b2ncon}}), if $n$ is greater than 18, the aforementioned distribution can approximate a normal distribution $\mathcal{N}(0.5n, 0.25n)$. 
We adopt this method in the design of a GRNG. 
For example, an implementation can use 
a 128-bit \emph{Linear Feedback Shift Register} (LFSR) for random bits generation and a \emph{Parallel Counter} (PC) to convert the number of 1's in the LFSR to a binary number\cite{andraka1998fpga}. As a simple method 
to generate uniformly distributed (pseudo) random numbers, an LFSR \cite{David:1980:TFS:1311064.1311138} implements a well-chosen \emph{linear feedback function}, allowing it to produce all binary combinations in a random order. 

When the depth of LFSR (i.e., number of registers) is large enough, e.g., 128, 
it costs 3000 years to cycle through when clocked at 80MHz\cite{andraka1998fpga}. 
We believe that the degree of randomness is high enough when the number of registers is large.
The LFSR implements the linear feedback function by 
shifting and updating the values in the \emph{taps} using XOR gates. 
An 8-bit LFSR is shown in Figure \ref{fig:exam} (a). 
The taps for the 8-bit linear feedback function are 4, 5, and 6. The LFSR uses fixed head location (register 1) and shifts its contents in each cycle. Registers at tap locations 
are updated with the XOR result of their left neighbor and the head. 
For each tap $t$:
\begin{equation}
\label{eq:lfsrop}
R(t) \leftarrow R(t+1)\;XOR\;R(1)
\end{equation}
where $R(t)$ denotes the value of $t$-th register in the LFSR, and $R(1)$ is the value of the head.

By using a PC, the output of LFSR can be converted to a binary number representing the number of 1's in LFSR. The PC can be implemented using adders in a tree structure. 
However, the hardware cost for larger inputs can be 
huge as a 127-input PC requires 120 full adders. 
This motivates the following RAM-based implementation.

\subsubsection{RAM-Based Linear Feedback Function}

Despite conceptually straightforward, 
the binomial distribution approximation method using LFSR and PC is 
not suitable for parallel implementation due to 
its high usage of registers and adders. The LFSR requires huge register resources. In addition, even though only taps are updated in each iteration (the numbers of taps are always 3 for 4-bit to 2048-bit LFSR \cite{ward2007table}), the PC needs to accumulate all bits of LFSR. 
Such large PC not only requires huge amount of hardware resources,
but also leads to extra computation latency. 

To overcome the challenge, 
we propose RLF, a compact {\em RAM-based Linear Feedback function} implementation,
which incurs lower cost but achieves the the same functionality as LFSR.
The design requires a much smaller PC 
that only calculates {\em the summation of tap values}.

Figure \ref{fig:exam} (b) illustrates the operations using
an 8-bit RAM-based Linear Feedback (RLF) logic. 
RLF logic stores the seed in fixed location while using 
{\em a self-accumulating indexer}, which changes in each cycle, to
track the head and taps to produce a pseudo shift operation. 
The updated results are fed back to the same locations. 
Compared to equation (\ref{eq:lfsrop}), for each tap $t$ in the 
linear feedback function, the corresponding operations on the RLF logic is:
\vspace{-0.4em}
\begin{equation}
\label{eq:rlfop}
x(h+t) \leftarrow x(h+t)\;XOR\;x(h)
\vspace{-0.4em}
\end{equation}
where $h$ is the current head location, and $x(i)$ is the value stored in the $i$-th entry of the vector (seed memory). 

As shown in Figure \ref{Fig:rlflogic}, the proposed RLF logic consists of four components: the seed memory (SeMem), the indexer, the updater and the buffer register. The SeMem stores the seeds using several blocks of 2-port RAM. The length of the SeMem is the size of the seeds, and the word width of the SeMem is the number of parallel RLF logics. 
The buffer register caches the values of taps and the head. 
The updater performs XOR based operations. 
The indexer stores the locations of the taps and the head, i.e. $h$ and $h+t$ for all taps in equation (\ref{eq:rlfop}), and increments them every cycle. 
This organization allows: 
{\em 1)} efficient hardware utilization for parallel operations,
because only one indexer is needed regardless of the number of RLF logics running in parallel;
{\em 2)} very small PC needed to calculate the seed summation,
because only the values of the taps, instead of all seeds, are outputted.    

Based on the basic structure of the proposed RLF logic,
we perform optimizations in two aspects:
{\em 1)} the quality of random numbers generated from the seeds, and 
{\em 2)} seed storage scheme. 
To explain the ideas, we consider a 255-bit RLF logic for 8-bit GRNG (each GRN uses 8-bit representation).
The taps for the 255-bit linear feedback function are 250, 252, and 253. According to equation (\ref{eq:lfsrop}), they need to be updated according to the following operations:
\begin{subequations}
\begin{gather}
x(h+250) \leftarrow x(h+250)\;\mathrm{XOR}\;x(h)\\
x(h+252) \leftarrow x(h+252)\;\mathrm{XOR}\;x(h)\\
x(h+253) \leftarrow x(h+253)\;\mathrm{XOR}\;x(h)
\end{gather}
\label{tap255}
\end{subequations}
The index $i$ is always less than or equal to 255, i.e., $i\leftarrow i-255$ if $i>255$. 
As the number of taps of the 255-bit linear feedback function is 3, the absolute difference between the output summations in two consecutive cycles cannot exceed three. This could affect the quality of random numbers produced, consequently undermining the performance of the BNN. To address this issue, the implemented linear feedback is modified to increase the number of taps by {\em combining two consecutive updates to create operations involving more bits}, which are:
\begin{subequations}
\begin{gather}
x(h+250) \leftarrow x(h+250)\;\mathrm{XOR}\;x(h)\\
x(h+251) \leftarrow x(h+251)\;\mathrm{XOR}\;x(h+1)\\
x(h+252) \leftarrow x(h+252)\;\mathrm{XOR}\;x(h)\\
x(h+253) \leftarrow (x(h+253)\;\mathrm{XOR}\;x(h))\;\mathrm{XOR}\;x(h+1)\\
x(h+254) \leftarrow x(h+254)\;\mathrm{XOR}\;x(h+1)
\end{gather}
\end{subequations}
By combining two consecutive updates into a single one, the maximum absolute difference between the outputs in two cycles is increased to five. Accordingly, the index increment per cycle is increased from one to two. The taps for the updated linear feedback function is from 250 to 254. Therefore, each update needs to read seven entries (the head, and second head, and all taps) and write five entries (the taps). The read/write bandwidth can be greatly reduced by optimizing the buffer register.

\begin{figure}[t]
\centering
\includegraphics[width = 3in]{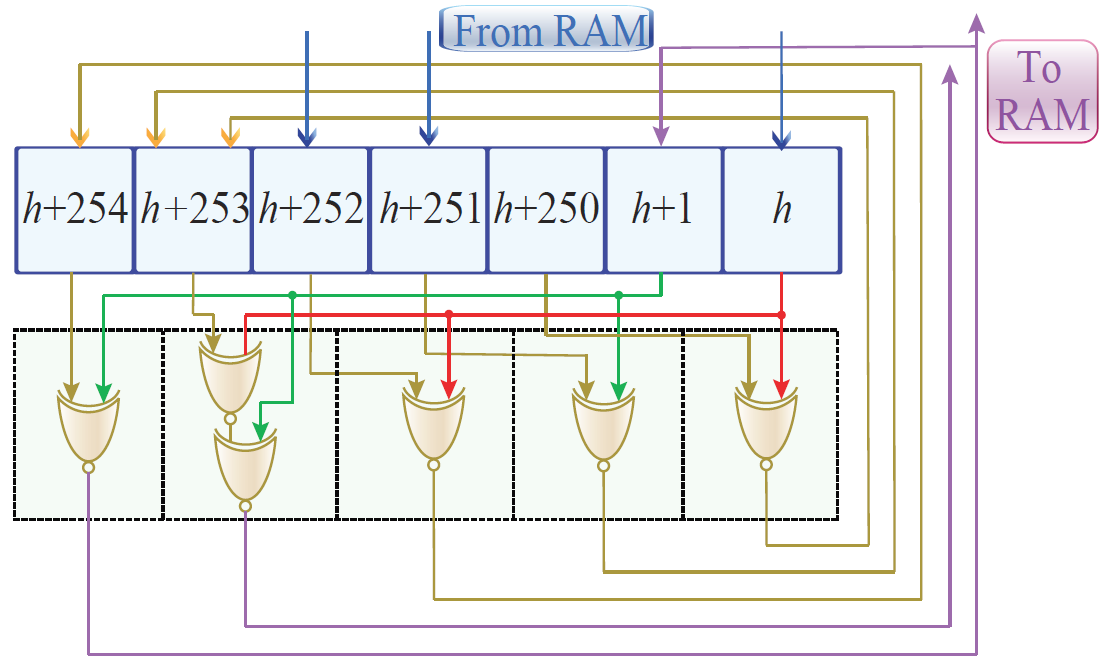}
\caption{Buffer register and updater for 8-bit RLF-GRNG}
\label{Fig:bff}
\end{figure}

As illustrated in Figure \ref{Fig:bff}, the buffer register and the updater are designed to minimize RAM access bandwidth. As the combined linear feedback function has five taps and two heads locations, the buffer register is 7-bit. Since the index is increased by 2 in each iteration/cycle, the left 5 bits, which represent the taps, all shift left by 2 bits when being updated; while the right 2 bits are read from the RAM. The updated values of the left 2 bits are then written back to the RAM. Given this circulant operation, the updated value of the leftmost bit ($h+254$) is the bit next to the head in the following iteration (as $mod(h+254+2, 255) = mod(h+1, 255)$). Therefore, aided by the buffer register and 2-port RAM, each iteration only requires 3 entries to read and 2 entries to write:\\ 
Read: $x(h)$, $x(h+250)$ and $x(h+251)$. \\
Write: $x(h+253)$ and $x(h+254)$.

\begin{figure}[!tb]
	\centering
  \begin{minipage}{1\columnwidth}
 	\centering
    	\includegraphics[width=2.8in]{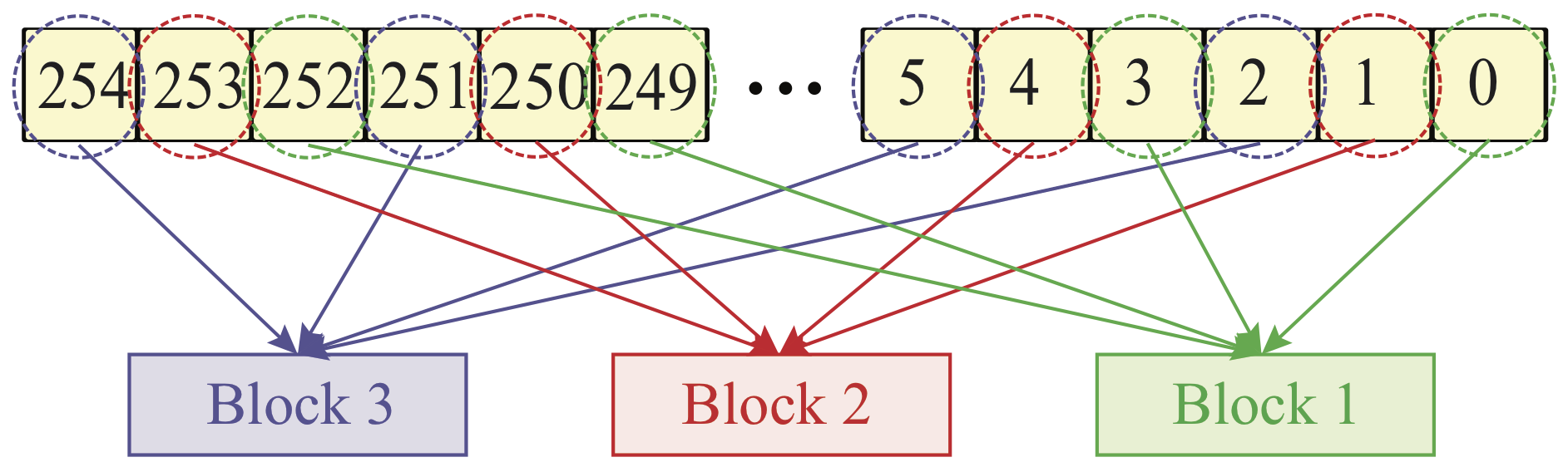}
        \captionof{subfigure}{Seed storing scheme}
   	\centering
    	\includegraphics[width=2.8in]{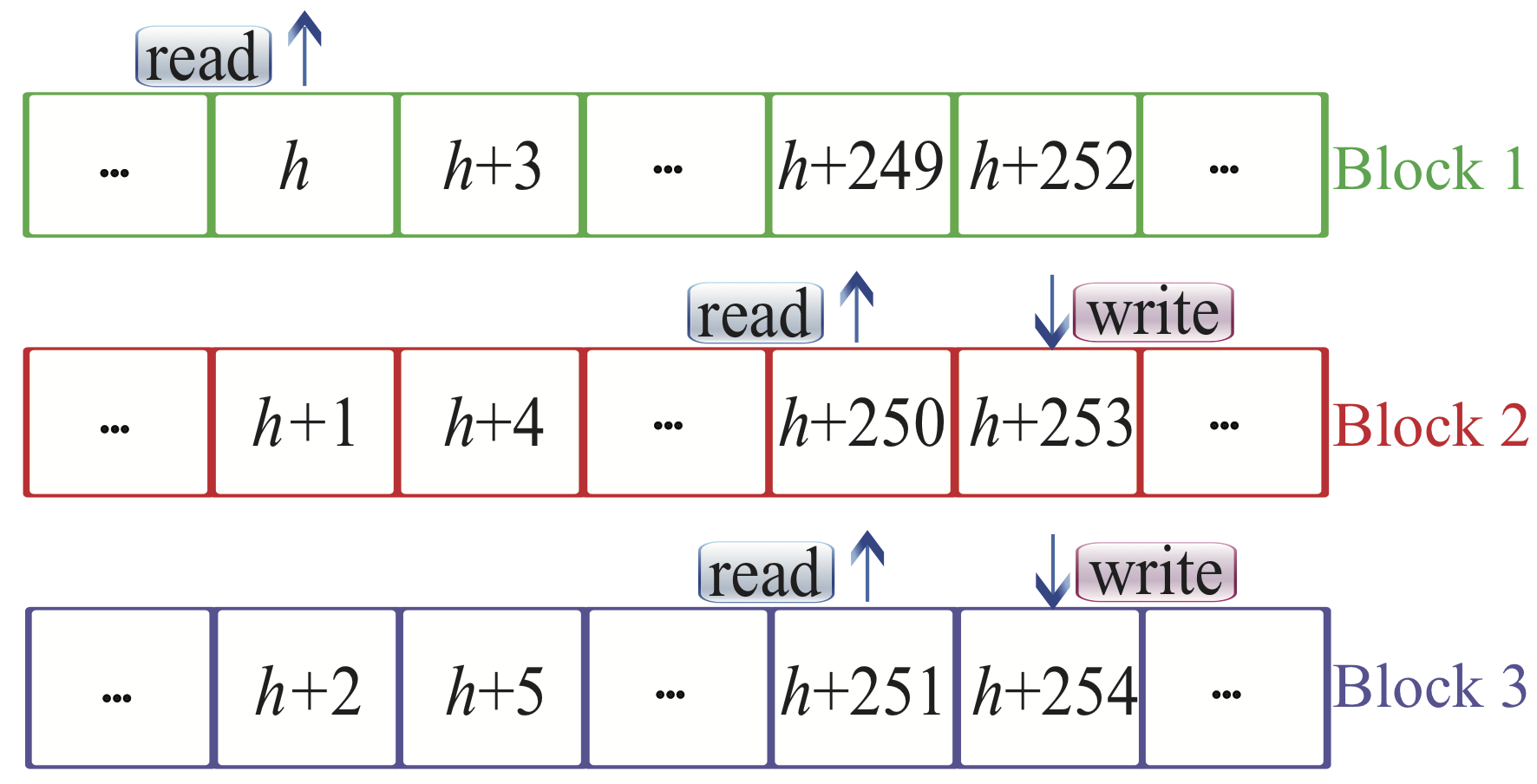}
       \captionof{subfigure}{RAM operations}
    \end{minipage}
    \caption{A 3-block RAM storing scheme for the 255-bit seed}
	\label{fig:ramblock}
\end{figure}

\begin{figure}[!tb]
\centering
\includegraphics[width = 3.2in]{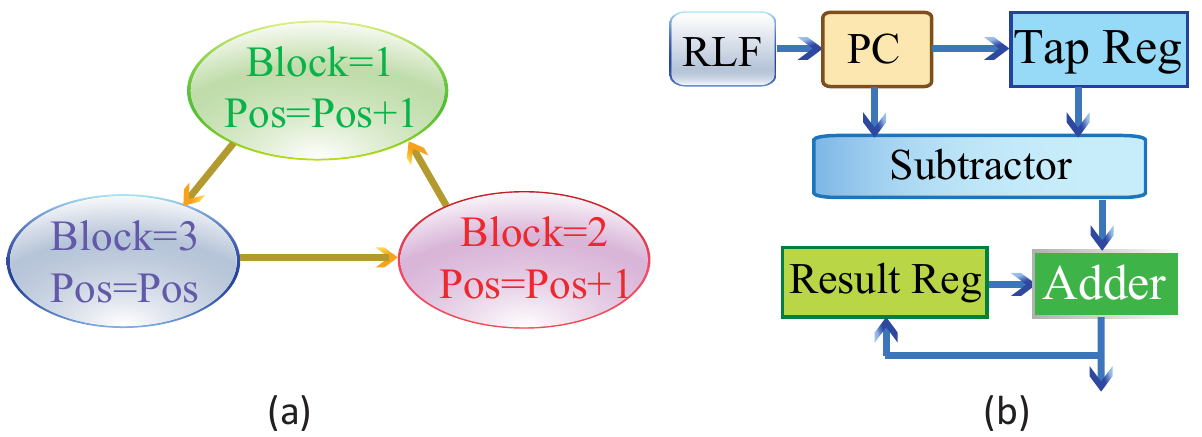}
\caption{(a) FSM for the indexer (b) Data Flow Graph of the RAM-based Linear Feedback GRNG (RLF-GRNG) }
\label{Fig:fsmind}
\end{figure}

To allow the above operations performed in one cycle on 2-port RAM, we propose a {\em 3-block RAM storing scheme} for the 255-bit seeds. As shown in Figure \ref{fig:ramblock}, the 255-bit seeds are separated evenly into three blocks based on the modulo operation results over three of their locations. Therefore, the five r/w operations on RAM can be performed in one cycle on three 2-port RAM blocks. The indexer, which stores and updates the locations for each tap and the head, can be implemented as a simple state machine shown in Figure \ref{Fig:fsmind} to track the block number (Block) and the relative position (Pos).  

\subsubsection{Overall RLF-GRNG Design}
The data flow graph of the proposed RAM-based Linear Feedback GRNG (RLF-GRNG) is shown in Figure \ref{Fig:fsmind}. The RLF logic, as discussed before, contains a SeMem for seeds storing, a buffer register for caching taps, an LF updater for updating taps, and an indexer to track tap locations. The RLF logic outputs a stream of updated taps to the PC, which accumulates the output stream. The previous tap summation is stored in the {\em tap register}. Their difference can be obtained from the subtractor. Finally, the difference is accumulated to the previous random number which is stored in the {\em result register} to produce a new output. Please note that, the initial value stored in the result register is the summation of the initial values stored in the SeMem. Therefore, this value can be {\em pre-calculated and stored in memory}. 

\begin{figure}[t]
\centering
\includegraphics[width = 3in]{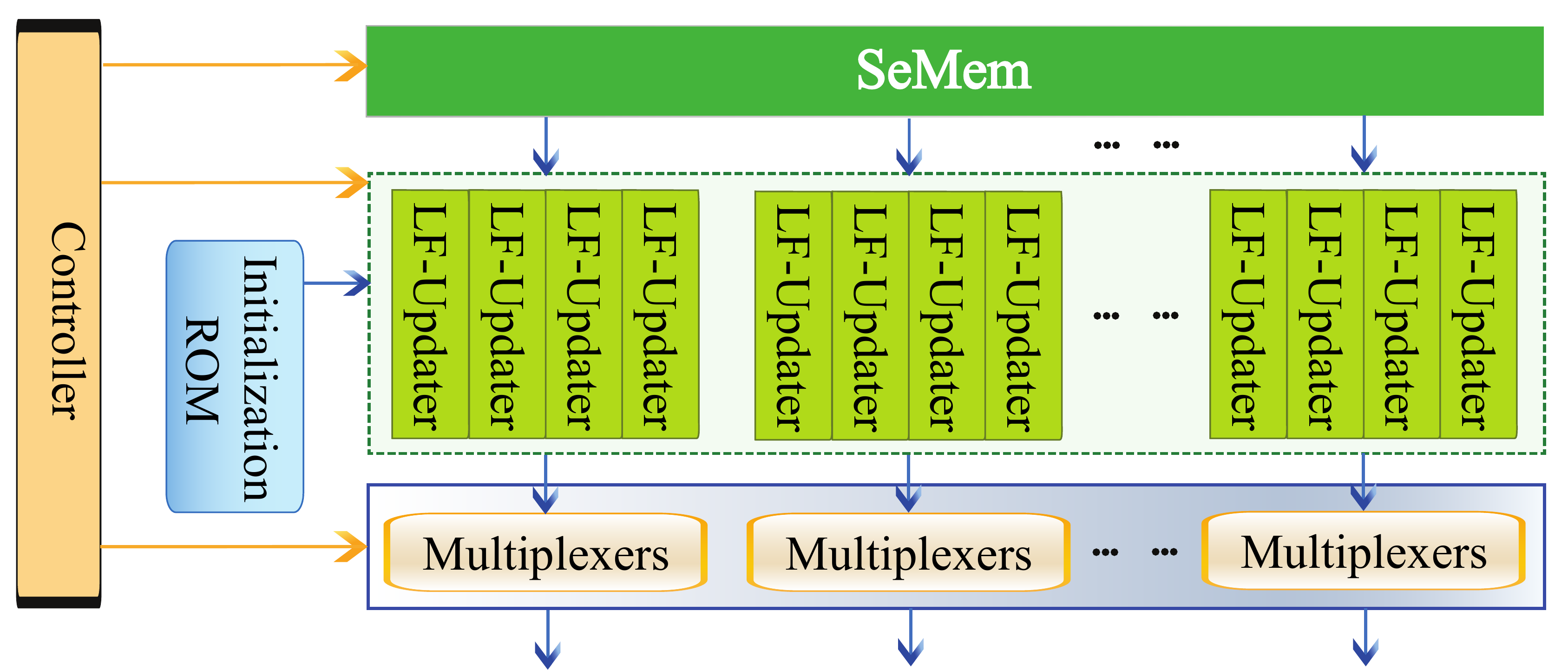}
\caption{Hardware block diagram of the 2-stage RLF-GRNG}
\label{Fig:ovr}
\end{figure}

Figure \ref{Fig:ovr} shows the block diagram of the parallel RLF-GRNG, designed for
the efficient implementation of BNNs. The Initialization ROM stores the initial summation results of the seeds for the GRNG. 
The SeMem stores all seeds for random number generations. 
To produce $m$ $n$-bit random numbers in parallel, the size of the SeMem is $2^{n}-1$ words, where each word is $m$-bit. 
Each bit of the word read from the RAM is propagated to an LF-updater 
for tap update and random number calculation. 
The updated taps are collected from all LF-updaters and formed into one word to be written back to the SeMem. The results generated by every four LF-updaters are selected sequentially to four outputs, with different orders through its multiplexer for enhanced randomness. All select signals are shared and generated by the controller. The controller also produces indices and memory access signals for the SeMem, as well as command signals for all LF-updaters. Finally, a set of individual Gaussian random numbers are produced in parallel.

\subsection{BNNs-Oriented Wallace GRNGs}
\subsubsection{The Wallace Method}
The Wallace method relies on the property that the linear combinations of Gaussian random numbers are still Gaussian distributed. The linear combinations are achieved through multiplying a vector of Gaussian random numbers $x$ by a Hadamard matrix $H$: $x' = H \times x$. Below is a typical Hadamard matrix:
\[H=\begin{bmatrix}
    -1     & \quad1 & \quad1 & \quad1 \\
    \quad1 & -1     & \quad1 & \quad1 \\
    -1     & -1     & \quad1 &     -1 \\
    -1     & -1     &     -1 & \quad1
\end{bmatrix}\]
Thus, the transformed random numbers are:
\begin{equation}
\begin{gathered}
x'[1] = t - x[1]; \quad  x'[2] = t - x[2]; \\
x'[3] = x[3] - t; \quad  x'[4] = x[4] - t; 
\end{gathered}
\label{eq:hadamard_transform}
\end{equation}
where $x'[1]-x'[4]$ are newly generated (pseudo) random numbers, $x[1]-x[4]$ are original random numbers fetched from the pool, $t=\frac{1}{2}(x[1]+x[2]+x[3]+x[4])$. This approach does not require multiplication operations. In Wallace algorithm, the original elements in $x$ are randomly chosen from the pool, and after multiple loops of transformations, the newly generated random numbers are written back to random positions in the pool to replace the original elements. In this way, the size of the pool keeps constant. However, the exact implementation in hardware costs another random number generator to produce the random positions, as well as extra delay to perform the multi-loop transformations. Besides, the initial pool should be large enough to ensure the randomness of the generated random numbers and the stability of $(\mu, \sigma)$. In Section \ref{Sec:BNNWallace}, we propose a {\em sharing and shifting scheme} that is proven to be capable of overcoming the downsides of the algorithm. 

\subsubsection{Hardware Design and Optimization of the Wallace Method}
\label{Sec:BNNWallace}

\begin{figure}[htbp]
\centering
\includegraphics[width=0.4\textwidth]{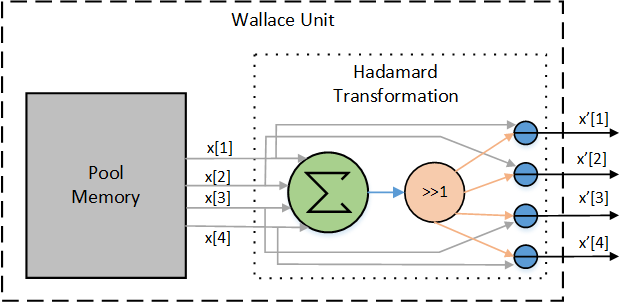}
\caption{Architecture of the Wallace Unit}
\label{Fig:wallace_unit}
\end{figure}

\begin{figure}[htbp]
\centering
\includegraphics[width=0.3\textwidth]{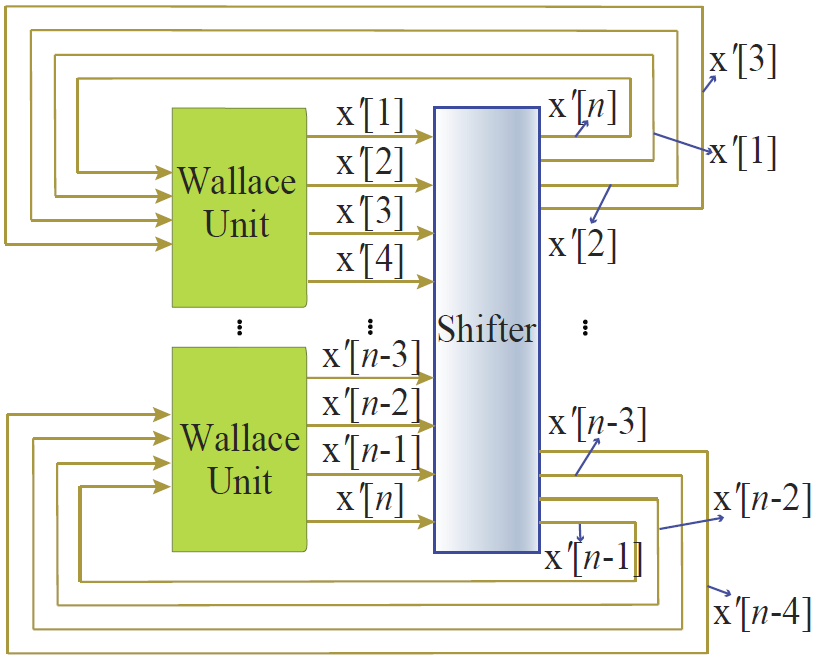}
\caption{Architecture of the BNNWallace-GRNG}
\label{Fig:wallace_architecture}
\end{figure}

Figure \ref{Fig:wallace_unit} illustrates the hardware design for performing one Hadamard transformation. The Pool Memory stores the pre-generated Gaussian random numbers, and four random numbers $x[1]-x[4]$ are fetched from the pool in every clock cycle. Then a summation is performed on the four random numbers, which is followed by a one-bit right-shifting to achieve the operation of dividing by 2. The output of the shifter is $t$ mentioned in equation \ref{eq:hadamard_transform}, then $t$ will participate in four parallel subtractions to generate new random numbers $x'[1]-x'[4]$. This module is named the \emph{Wallace Unit}, 
and it is the basic component in our proposed \emph{BNN-oriented Wallace GRNG (BNNWallace GRNG)}. Even though a larger Hadamard matrix can be composed of small Hadamard matrices as $H'=\bigl(\begin{smallmatrix} H & H  \\ H & -H \end{smallmatrix}\bigr)$, using the larger one implies that each new random number $x'[i]$ is generated by the computations involving all original random numbers. For instance, if we perform a Hadamard transformation using a $8 \times 8$ matrix, $x'[1]=\sum_{i=1}^n x[i] - (x[1]+x[5])$. 
Therefore, more hardware resources and longer clock cycle are needed. 
Based on the above analysis, we implement the Wallace Unit using the $4 \times 4$ Hadamard matrix. 

As introduced before, large initial pool and multi-loop transforms are required to guarantee 
that the generated random numbers are highly random, less correlated, 
and stably distributed. These requirements or downsides of Wallace method are substantially magnified when it is realized on hardware due to the limited resources.
We introduce a {\em sharing and shifting-based optimization} 
to overcome these downsides. The key insight is:
to make small pools of Wallace Units work as a whole, so the memory requirement on each pool is roughly divided by $N$ if there are $N$ Wallace Units. This is achieved by shifting the generated random numbers by one number before they are written back to the pool, and Figure \ref{Fig:wallace_architecture} illustrates the scheme. Due to the shifting, the written back random numbers are partially generated by other Wallace Units. This strategy can increase the randomness and decrease the correlations among the random numbers. Besides, 
by shifting, the generated random numbers flow through all Wallace Units, so all small pools constitute a large pool. 
As a result, the stability of $(\mu, \sigma)$ is guaranteed. 
The experimental results in Section \ref{results_grngs} demonstrate 
the effectiveness of BNNWallace-GRNG.

\section{FPGA-Based Implementation of VIBNN}
To implement VIBNN architecture 
on resource-limited FPGAs, we explore a series of optimization strategies 
at three levels: 
{\em the arithmetic unit-level, the PE-level, and the system-level}, 
in addition to the optimized GRNGs. 

\subsection{PE Design}
\begin{figure}[htbp]
\centering
\includegraphics[width=0.4\textwidth]{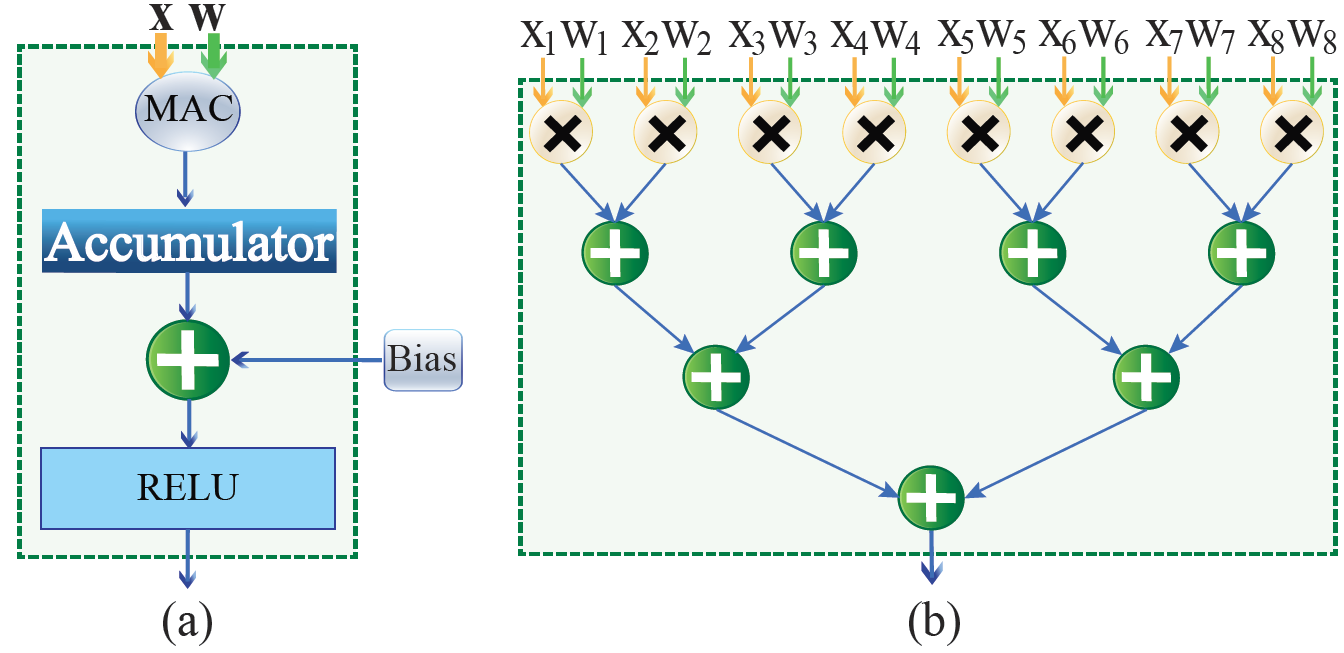}
\caption{(a) PE Architecture (b) MAC Architecture}
\label{Fig:pe_architecture}
\end{figure}

In VIBNN, a PE works as a neuron with architecture illustrated in Figure \ref{Fig:pe_architecture}-(a). The Multiplication and Accumulation (MAC) unit with architecture shown in Figure \ref{Fig:pe_architecture}-(b) calculates the dot-product of its input features and weights through multipliers and an adder tree. Typically, a neuron in FNNs has hundreds of inputs, but it is infeasible for a PE due to the limited resources on FPGAs. 
A large number of neurons per PE will decrease the number of PEs that
can be implemented on a FPGA,
leading to a rigid system implementation 
and limited optimization space. 
Therefore, we choose to implement a PE with a reasonable number of inputs and use it in a time-multiplexed manner. Inside a PE, we use an accumulator to accumulate the partial dot-products. After some iterations, the accumulated value will be sent to an adder to add a bias up. Next, the biased dot-product is fed into the \emph{Rectifier Linear Unit (ReLU)} so that the activation result is obtained.

\subsection{Bit-Length Optimization}
Since the arithmetic units are implemented in look-up tables (LUTs) on FPGAs, there is no much optimization space on the structures and implementation principles of those units. The most straightforward and feasible method is to optimize the bit-length of operands, e.g., the input and output size of the arithmetic units. Despite the previous works such as \cite{han2015deep,gysel2016hardware,venkatesh2017accelerating} leveraging 
this technique, it is not trivial to investigate and apply this technique in VIBNN, 
because the introduced random noises may lead to a different conclusion.
We select 16-bit as the starting point, which is widely adopted in hardware neural network implementations, 
then a binary search is used 
to figure out the smallest required bit-length that can guarantee acceptable accuracy.
 Optimization results are reported in Section \ref{results_bitlength}. 

\subsection{Weight Generator}
\begin{figure}[htbp]
\centering
\includegraphics[width=0.35\textwidth]{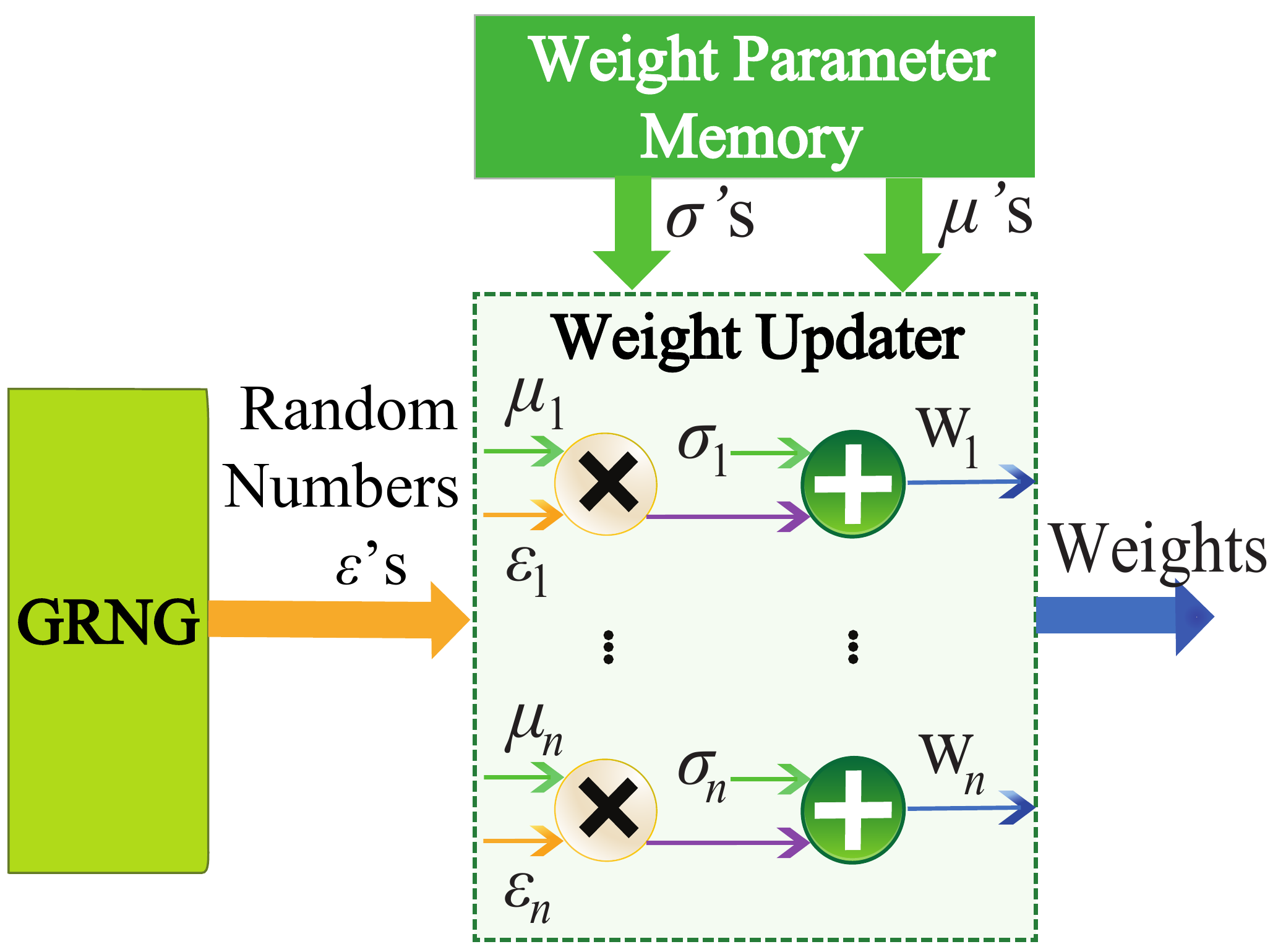}
\caption{Architecture of the Weight Generator}
\label{Fig:weight_generator}
\end{figure}

As shown in Figure \ref{Fig:weight_generator}, the Weight Generator consists of a GRNG, a Weight Updater, and an on-chip Weight Parameter Memory. The Weight Updater receives random numbers sampled by GRNG, then generates a weight sample according to the corresponding variational parameters read from the memory. It applies the variational parameters ($\mu$ and $\sigma$) to the Gaussian numbers using multipliers and adders. 

\subsection{Joint Optimization of PE Size/Number and Memory Access}
The throughput of a neural network accelerator is mainly determined by computation
and communication~\cite{zhang2015optimizing}. In FNNs, the computations can be reduced by applying optimizations that reduce links between neurons, such as weight pruning \cite{han2015deep}. 
In our work, we focus on a different aspect. 
Specifically, from the computation aspect, the throughput can be improved 
by increasing computation parallelism, 
in the form of increasing PE size or increasing the number of PEs. 
From the communication aspect, the throughput improvements come from reducing the memory traffic. The key problem is that, 
the computation parallelism and memory traffic are not completely independent factors, 
so we need to jointly consider the parallelism and memory traffic to obtain the best throughput performance.

On-chip memories are divided into two categories according to different usage purposes:
{\em 1)} IFMem (input feature memory): the memory that holds input features as well as activation outputs; and 
{\em 2)} WPMem (weight parameters memory), the memory that holds weight parameters $(\mu, \sigma)$'s. The optimization techniques for reducing memory accesses to 
IFMems and WPMems need to be considered separately.

\begin{figure}[htbp]
\centering
\includegraphics[width=0.5\textwidth]{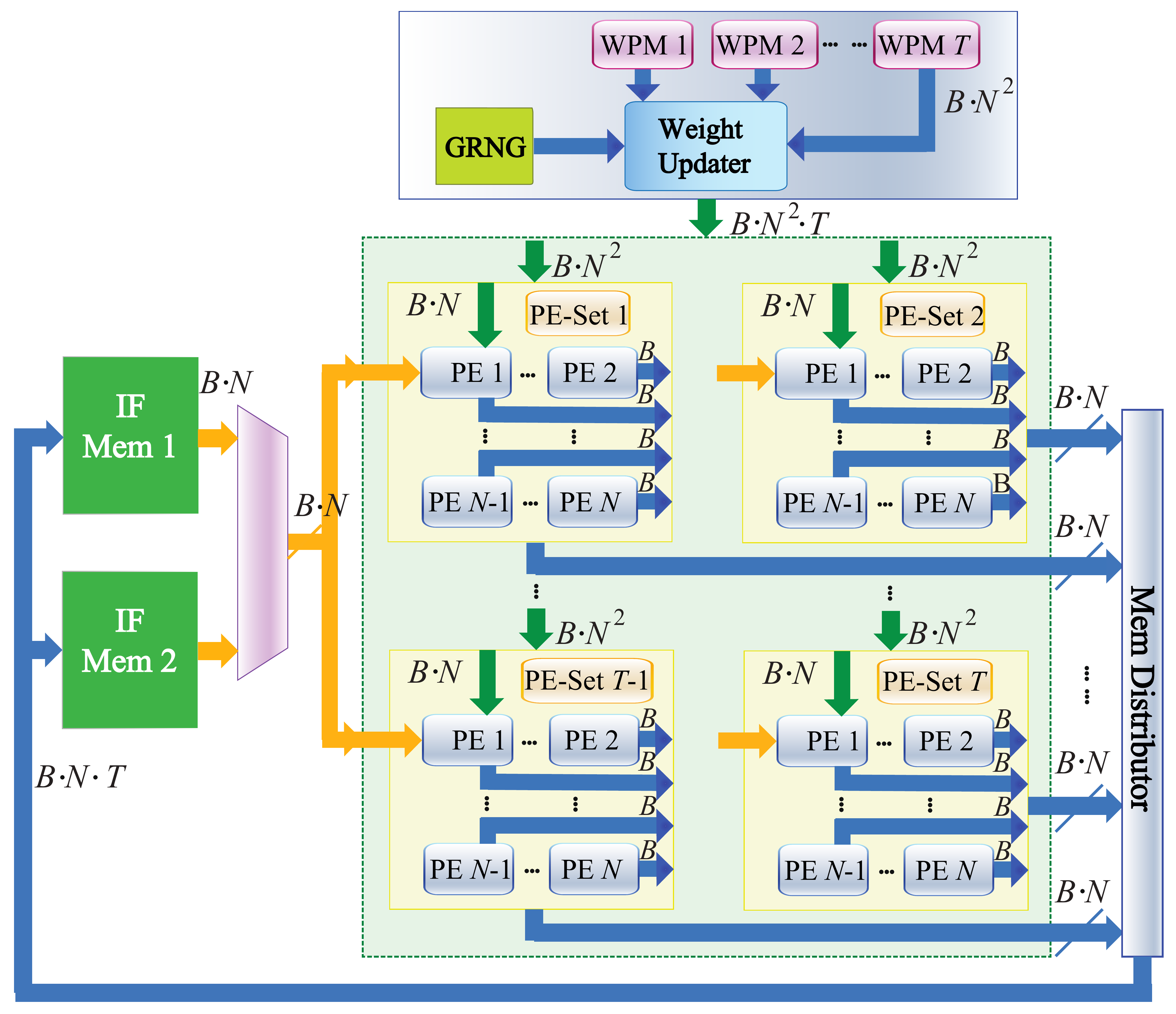}
\caption{Joint Optimization Architecture}
\label{Fig:joint_optimization}
\end{figure}

\subsubsection{Reducing IFMem Accesses}
First, our hardware realized BNNs are aimed at FNN models and one PE corresponds to one neuron, any two PEs will have the same input features in a certain cycle. Therefore, the word size of the IFMem should be $B \times N$ to avoid extra delays:
all required input features can be sent to PEs at each access to IFMem. As illustrated in Figure \ref{Fig:joint_optimization}, in our design, the whole network is constructed on the basis of PE-sets and each set has $S = N$ PEs. Thus the outputs of a PE set correspond to a word to be written back to the IFMem, and all the outputs from different PE sets are buffered in the Memory Distributor. To illustrate how the model works and its advantages, we define three 
additional parameters: 
{\em 1)} $T$: the number of PE sets; 
{\em 2)} $MinIn$: the minimum input size of a neuron in the target neural network to be realized; 
{\em 3)} $MaxWS$: the maximum allowable word size of an on-chip memory. Their relations are derived as: 
\begin{subequations}
\begin{gather}
T \times S < ceil(\frac{MinIn}{N})
\label{eq:pesize_cal1}\\
B \times N <= MaxWS
\label{eq:pesize_cal2}\\
S = N
\label{eq:pesize_cal3}\\
M = T \times S
\label{eq:pesize_cal4}
\end{gather}
\end{subequations}
where function $ceil(x)$ takes the ceiling of variable $x$, $B$ is the bit-length of an operand, $N$ is the input size of a PE, and $M$ is the total number of PEs.

\begin{figure}[b]
\centering
\includegraphics[width = 3.1in]{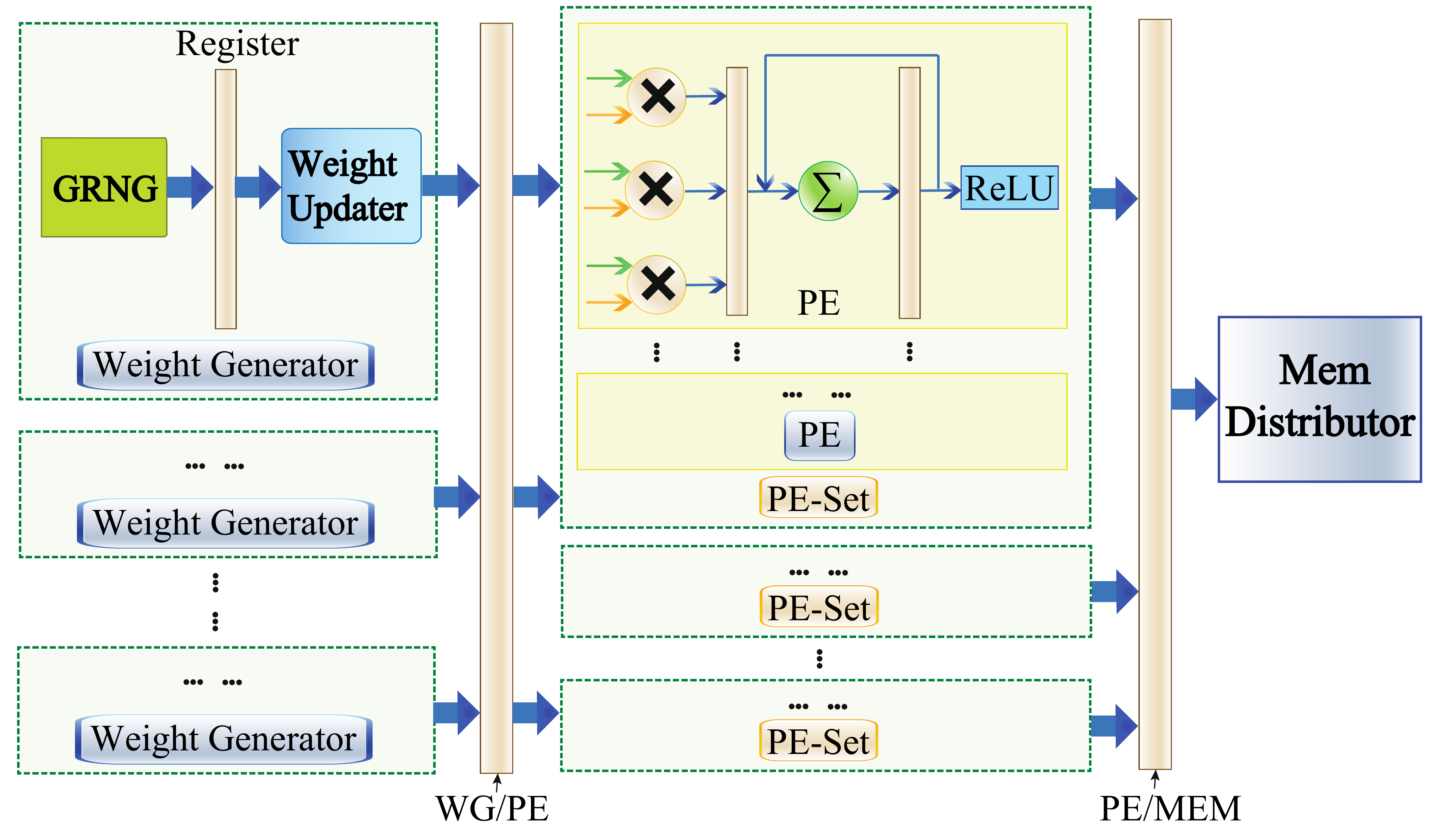}
\caption{Pipeline structure of the proposed BNN}
\label{Fig:pip}
\end{figure}

The advantages of the design are three-fold:
{\em 1)} scalability: we can easily change the PE size $N$ and PE sets number $T$ to
adopt to neural networks with various size, according to equation (\ref{eq:pesize_cal1}); 
{\em 2)} portability: we can change the PE size $N$, PE sets number $T$, 
and even bit-length $B$ to deploy the design on various FPGA platforms, according to equations (\ref{eq:pesize_cal1}) and (\ref{eq:pesize_cal2}); 
{\em 3)} access efficiency, equations (\ref{eq:pesize_cal1})-(\ref{eq:pesize_cal4}) guarantee that all the required inputs of PEs can be read at one access to the IFMem, and the buffered activation results will be written back to the IFMem before the next set of results coming into the Memory Distributor.

Besides, we use two IFMems alternatively to avoid any latent read\&write conflicts. For instance, at a certain layer, if IFMem1 is used to access input features, the activation results that are also input features for next layer will be written into IFMem2. Then at next layer, we switch the roles of IFMem1 and IFMem2. 

\subsubsection{Reducing WPMem Accesses}
Since each input among the whole network has an individual weight, to read all weight parameters required at a certain clock cycle, the word size should be $B \times N \times S \times T$.
Here, we only consider $\mu$'s for simplicity, and the WPMems for storing parameters $\sigma$'s are just duplicated. 
However, the maximum allowable word size $MaxWS$ restricts $N$, $S$, and $T$ to dramatically small values and the computation parallelism is hence restricted. 
Therefore, the optimization schemes for reducing IFMem accesses are not appropriate for WPMems. In VIBNN, we propose to use multiple distributed WPMems so that each WPMem corresponds to a PE set for maintaining a structural architecture and the ease of control. In this way, the bandwidth for a WPMem is $B \times N \times S$. Therefore, equations (\ref{eq:pesize_cal1})-(\ref{eq:pesize_cal4}) need to be rewritten as (\ref{eq:pesize_cal}). 
\begin{subequations}
\begin{gather}
T \times S < ceil(\frac{MinIn}{N})\\
B \times N \times S <= MaxWS\\
S = N\\
M = T \times S
\end{gather}
\label{eq:pesize_cal}
\end{subequations}
where function $ceil(x)$ takes the ceiling of variable $x$, $T$ is the number of PE sets, $S$ is the number of PEs in a set, $B$ is the bit-length of an operand, $N$ is the input size of a PE, and $M$ is the total number of PEs.

\subsection{Deep Pipelining}
As shown in Figure \ref{Fig:pip}, the proposed design implements a two-tier pipeline structure. The first tier is placed between weight generator and PE, which holds the sampled weights. The second tier is inserted inside the weight updater and the PE to further reduce system clock period. We insert DFFs between the GRNG and the weight updater. Each PE unit is optimized as a three-stage pipeline. The first stage calculates all multiplications; the next stage accumulates all internal products from previous stage. The ReLU is performed at the final stage.    

\section{Experimental Results and Analysis}
In this section, we present the experimental results regarding following respects: (i) the effectiveness and hardware cost of the proposed RLF-GRNG and BNNWallace-GRNG; (ii) the training characteristics of both FNNs and BNNs are compared when they are trained with small data; (iii) the bit-length optimization; (iv) the hardware resource utilization of the proposed VIBNNs; (v) the accuracy comparison between FNNs (software), BNNs (software), and our VIBNN (hardware) when they are applied on image classification task using MNIST dataset \cite{lecun2010mnist}, the Parkinson Speech Dataset\cite{6451090}, the Diabetics Retinopathy Debrecen Dataset\cite{Antal:2014:ESA:2611848.2612138}, the Thoracic Surgery Dataset\cite{zieba2013boosted} and the TOX21 Dataset\cite{attene2013tox21}.

\subsection{Performance of Proposed GRNGs}
\label{results_grngs}

The performance and hardware cost of the proposed GRNGs are tested and presented in this section. And also, their advantages and disadvantages are listed and compared according to the experimental results.

First, we implement an RLF-GRNG with 255-bit SeMem and a BNNWallace-GRNG with 8 Wallace Units, in which each has a pool of 256 initial random numbers. Then their performances in terms of randomness and stability are tested and compared with software-based Wallace method with various initial pool sizes. In addition, to illustrate the effectiveness of the sharing and shifting scheme of the proposed BNNWallace-GRNG, a hardware realized  Wallace Method with neither sharing and shifting scheme nor multi-loop transformations (Wallace-NSS) is implemented and compared.  
\begin{table}\small
\caption{Stability errors to $(\mu, \sigma)=(0,1)$ of Various Wallace Designs}
\label{tbl_wallace_stability}
\centering
\begin{tabular}{c|c|c}
\hline
GRNG Designs          &$\mu$ Errors  &$\sigma$ Errors  \\\hline
Software 256 Pool Size  &0.0012         &0.3050 \\
Software 1024 Pool Size &0.0010         &0.0850 \\
Software 4096 Pool Size &0.0004         &0.0145 \\
Hardware Wallace NSS    &0.0013         &0.4660 \\
BNNWallace-GRNG         &0.0006         &0.0038\\
RLF-GRNG                &0.0006         &0.0074\\
\hline
\end{tabular}
\end{table}

The results regarding the stability is shown in Table \ref{tbl_wallace_stability}. The distribution of the initial pool is sampled from the standard normal distribution, and the $\mu$ Errors and $\sigma$ Errors are the absolute errors between the generated distributions and the standard normal distribution. From the table, it can be observed that the size of the initial pool significantly affects the stability of the generated distributions. Results suggest that the proposed BNNWallace-GRNG and RLG-GRNG are comparable with the software method with the pool size 4096. Furthermore, both methods can generate multiple random numbers in parallel, which is a big improvement compared to the software method that can only generate 4 random numbers at one time. The memory savings of BNNWallace-GRNG is 2X, which can be further improved by sharing more Wallace Units because the pool size of each Wallace Unit can be further reduced when there are more units participating in the sharing and shifting. 

\begin{table}[t]\small
\centering
\caption{Hardware Utilization and Performance Comparison between RLF-GRNG and Wallace-based GRNG for 64 Parallel Gaussian Random Number Generation Task}
\label{tbl:hcomp}
\resizebox{\columnwidth}{!}{
\begin{tabular}{c|c|c}
\hline
\bf{Type} & \bf{RLF-GRNG} & \bf{BNNWallace-GRNG}\\\hline
Total ALMs & 831/113,560 ( $<1\%$ ) & 401/113,560 ( $<1\%$ ) \\
Total Registers & 1780 & 1166 \\
Total Block Memory Bits & 16,384/12,492,800 ( $<1\%$ ) & 1,048,576/12,492,800 ( $8\%$ ) \\
Total RAM Blocks & 3/1220 ( $<1\%$ ) & 103/1220 ( $8\%$ ) \\
Power(mW) & 528.69 & 560.25 \\
Clock Frequency (MHz) & 212.95 & 117.63 \\
\hline
	\end{tabular}
}
\end{table}

\begin{table}[h]\small
	\centering
	\caption{RLF-GRNG and BNNWallace-GRNG Comparison}
	\label{tbl:comp}
\resizebox{\columnwidth}{!}{
	\begin{tabular}{c|c|c}
    \hline
	\bf{} & \bf{RLF-GRNG} & \bf{BNNWallace-GRNG}\\\hline
	\bf{Advantages} & \bf{low memory usage} & \bf{Adjustable distribution} \\
    \bf{} & \bf{high frequency} &\bf{high scalability} \\
    \bf{} & \bf{high power efficiency} &\bf{low ALM and register usage} \\
    \hline
    \bf{Disadvantages} &\bf{low scalability} &\bf{high latency} \\
\hline
	\end{tabular}
}
\end{table}

\begin{figure}
\centering
\includegraphics[width=0.46\textwidth]{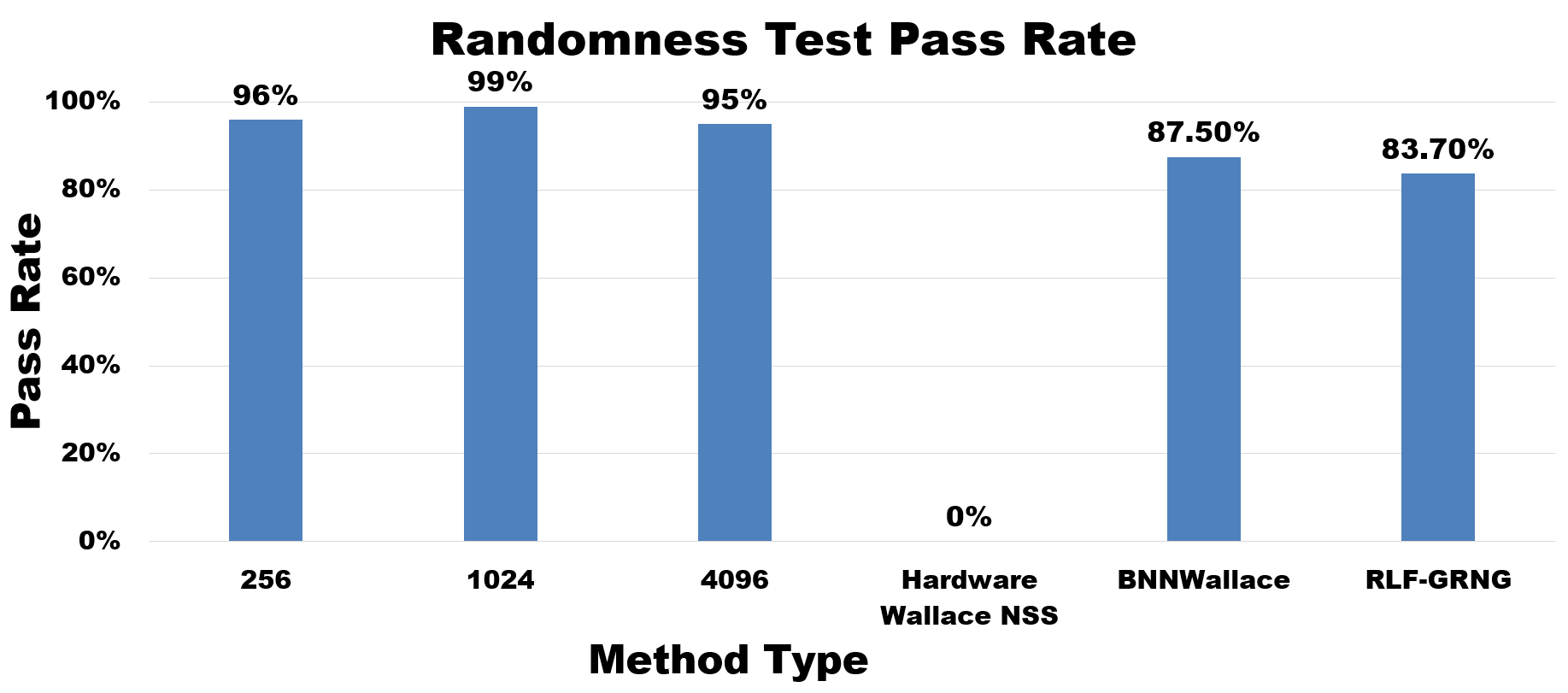}
\caption{Randomness tests pass rate of software-based Wallace method and BNNWallace with various pool sizes}
\label{Fig:runstest_wallace}
\end{figure}

In addition to the stability test, we also perform the randomness test on RLF-GRNG and various Wallace methods, which are shown in Figure \ref{Fig:runstest_wallace}. Each method is used to generate 100,000 random numbers and tested using the \emph{runstest} function in Matlab. The same test is repeated for 1000 times and the rates of passed tests are collected. It can be concluded that the pool size has insignificant impact on the randomness of the newly generated random numbers. However, which is worth noting is that our proposed BNNWallace-GRNG is comparable with the software-based methods, while the hardware version that has neither sharing and shifting scheme nor multi-loop transforms fails to pass any randomness test. This result verifies the effectiveness of our proposed GRNGs again.

Table \ref{tbl:hcomp} presents hardware performance comparison for RLF-GRNG and BNNWallace-GRNG. Both yield highly efficient hardware usage for 64 parallel Gaussian random number generating task. However, the RLF-GRNG utilizes less memory while the BNNWallace-GRNG requires less ALM resources. Given its shallow PC to accumulate taps and less complex computations, the RLF-GRNG is capable to operate at a higher frequency. 

As presented in Table \ref{tbl:comp}, both RLF-GRNG and BNNWallace-GRNG have certain limitations. The RLF-GRNG offers relatively limited generality as different size requires individual optimization.In addition, the fact that the RAM width required for RLF-GRNG is exponential to the bit length restricts its application to high bit-length Gaussian variable sampling. However, for 8-bit applications, RLF-GRNG is a highly memory efficient and  fast solution. The RLF-GRNG is much more power efficient than the BNNWallace-GRNG as its power consumption is the same as the BNNWallace-GRNG while operating at a much higher frequency. For parallel random number generation, RLF-GRNG performs better in terms of hardware resource utilization given the limited size of SeMem to implement one extra output.   

\subsection{Small Data}
\begin{figure}[t]
\centering
\includegraphics[width = 3.2in]{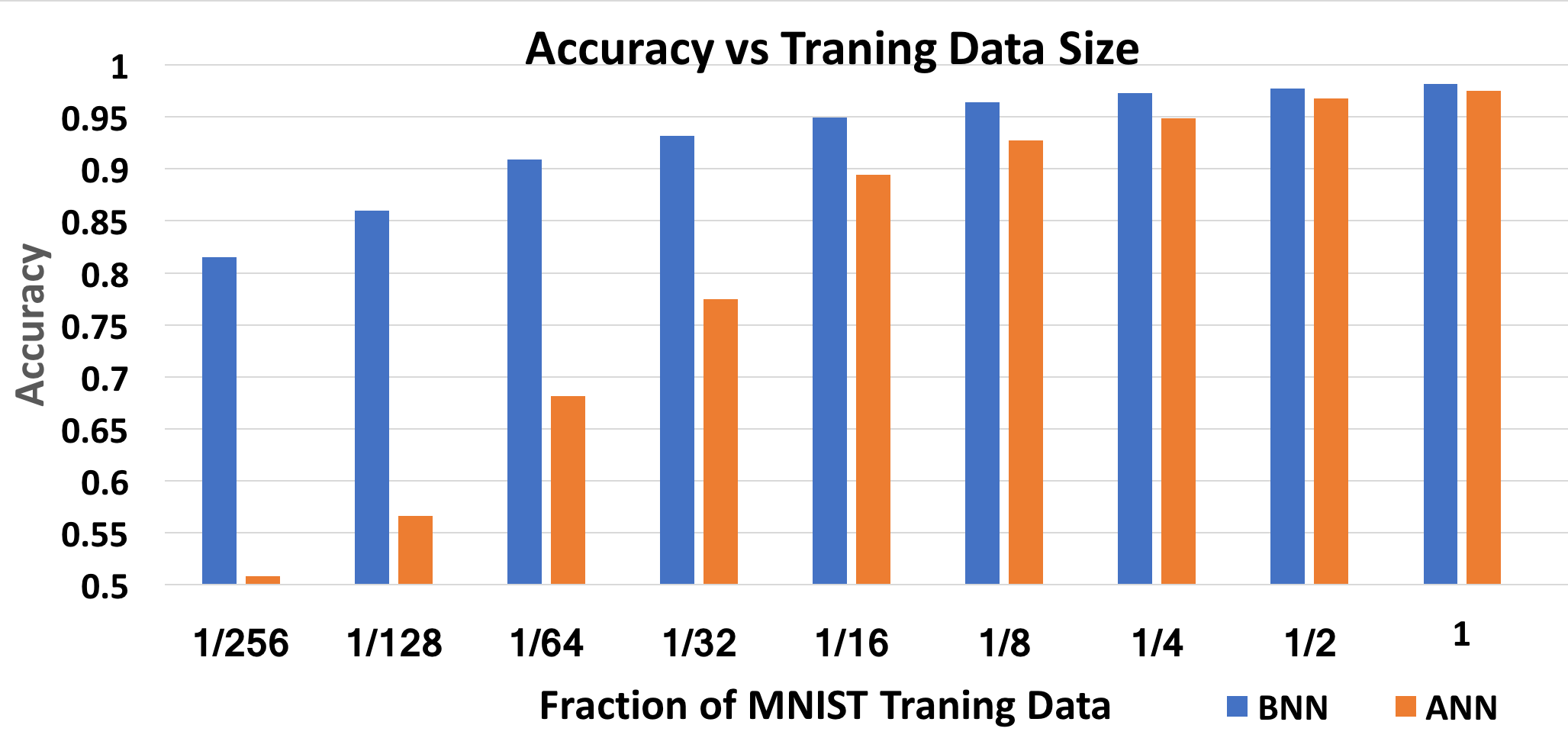}
\caption{Accuracy comparison between FNN and BNN with a fraction of training data}
\label{Fig:FracAcc}
\end{figure}

\begin{figure}[t]
\centering
\includegraphics[width = 3.2in]{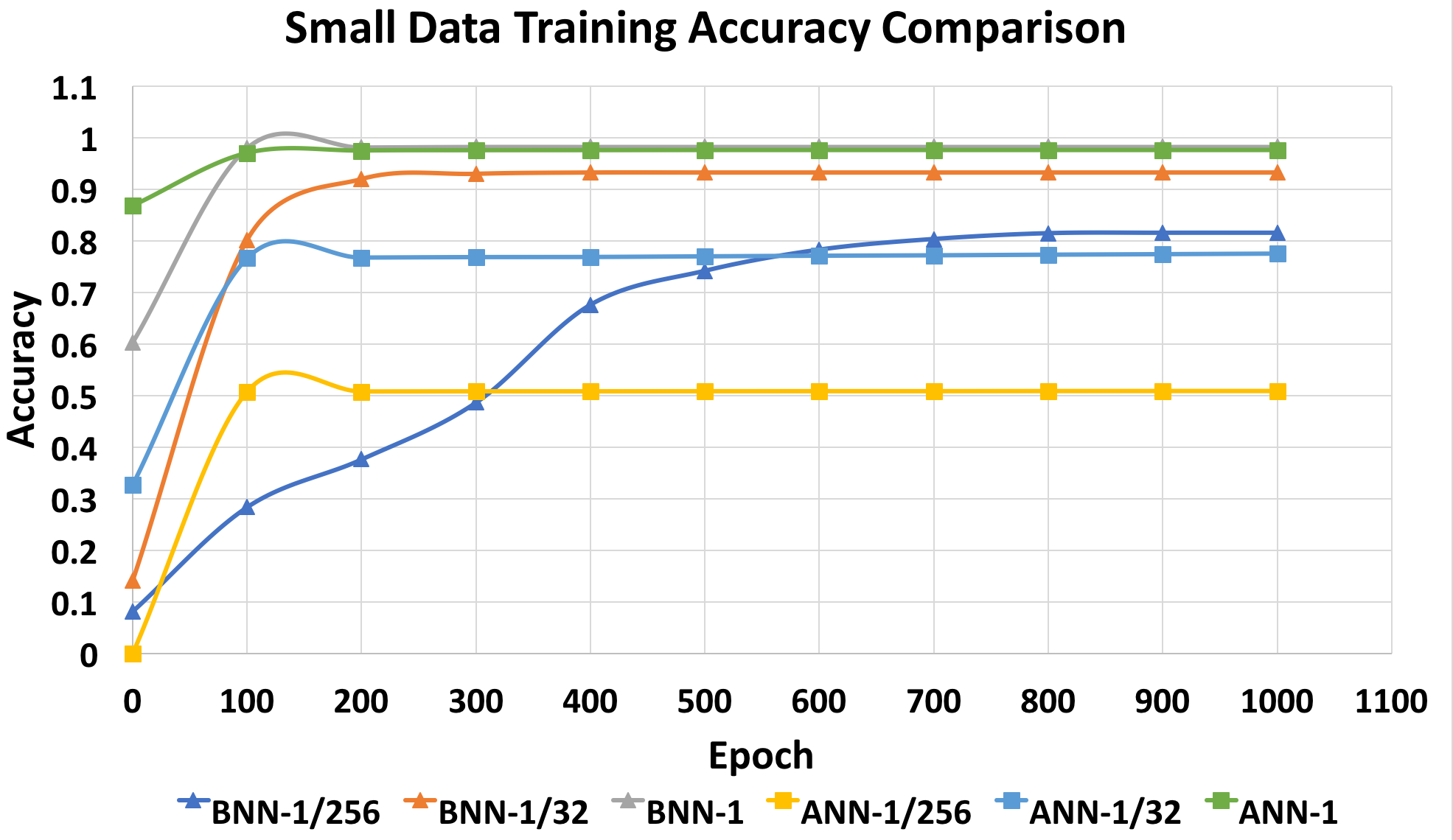}
\caption{Training accuracy comparison between FNN and BNN with a fraction of training data}
\label{Fig:smalltrain}
\end{figure}

To demonstrate the performance on small data, we test both BNN and FNN on the MNIST dataset\cite{lecun2010mnist}. The MNIST dataset contains 60K images for training and 10K images for testing. Each image is a $28\times28$ pixel gray-scale bitmap of hand written digit. The task is to classify the images into ten classes (from "0" to "9"). The BNN implemented has 784 inputs, 10 outputs, and 2 hidden layers both with 200 neurons. All layers are fully connected ($784 - 200 - 200  - 10$). We randomly choose a fraction of the training data to train a BNN and an FNN with same structure and test their performance. Figure \ref{Fig:FracAcc} shows accuracy comparison between FNN and BNN when training with part of the data from $1/256$ to entire training data set, and suggest that BNN performances much better as training data size shrinks. Detailed converge performance is shown in Figure \ref{Fig:smalltrain}.

\subsection{Bit-Length Optimization Results}
\label{results_bitlength}
\begin{figure}[t]
\centering
\includegraphics[width=0.46\textwidth]{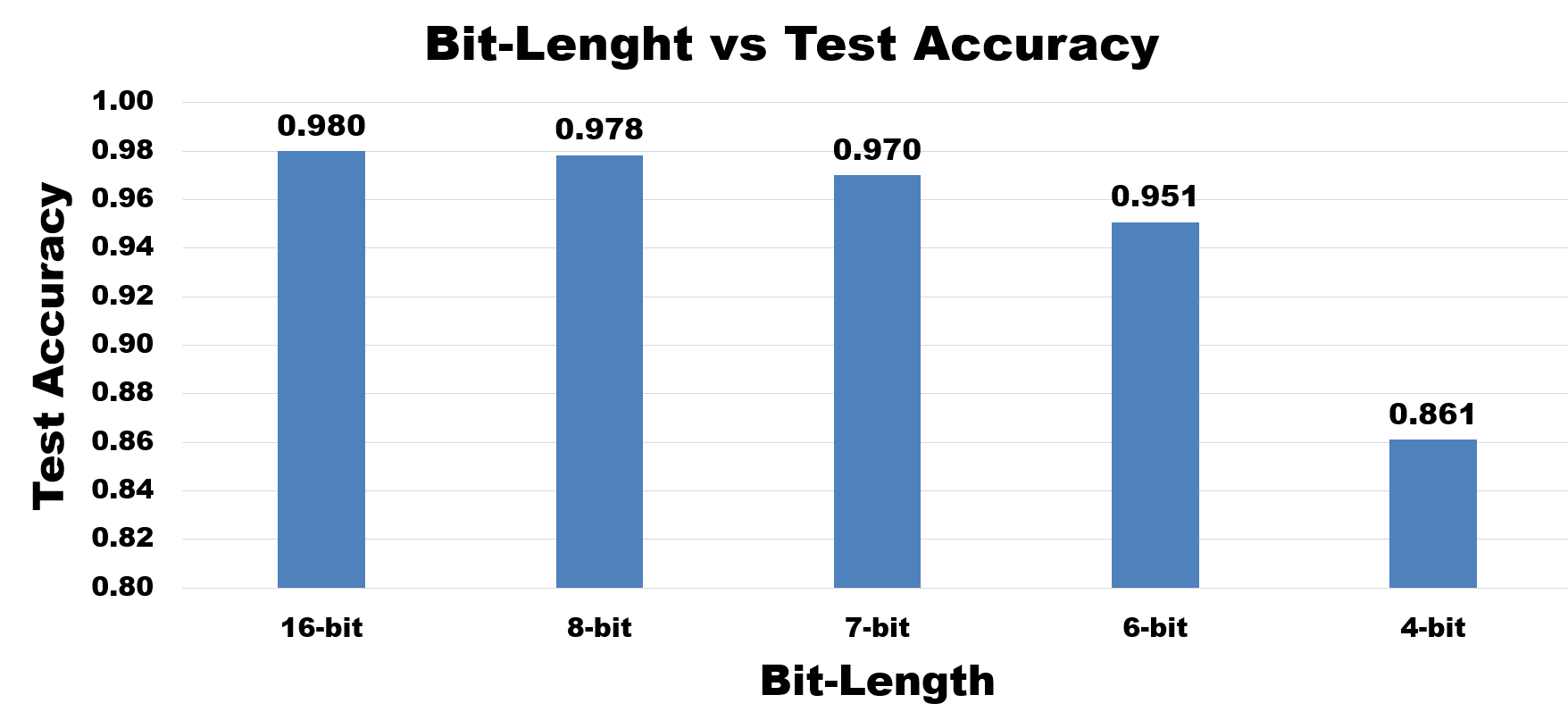}
\caption{Bit-Length vs Test Accuracy}
\label{Fig:bitlength_accuracy}
\end{figure}

According to our experiments on software, the BNNs can achieve 98.1\% test accuracy on image classification task using MNIST dataset, so we set 97.5\% as the threshold accuracy in selecting a proper bit-length. Figure \ref{Fig:bitlength_accuracy} shows the test accuracy when the bit-length is set to different values, and we can observe that 8-bit is the smallest bit-length that can reach the accuracy threshold. 

\subsection{Hardware Resources Usage}
We implement the proposed design of VIBNN on an Altera Cyclone V FPGA (Model Number 5CGTFD9E5F35C7). We opt for a 16 PE-set parallel design, each has eight 8-input PEs. The resource utilization results are shown in Table \ref{tbl_network_fpga}. It can be observed that both the two VIBNNs have high throughput and energy efficiency, and the RLF-based VIBNN provides even higher energy efficiency, while the BNNWallace-based VIBNN has higher design flexibility as explained in Section \ref{results_grngs}.

To demonstrate the efficiency of the proposed design, we compared the proposed implementation against software implementations on GPU(Nvidia GTX 1070) and CPU(Intel i7-6700k) on the MNIST dataset. As shown in Table \ref{tbl:perfcon}, the proposed design is $283\times$ more energy efficient compared with an Nvidia GTX 1070 and $458\times$ more energy efficient compared with an Intel i7-6700k processor. 

\begin{table}[t]
\centering
\caption{Summary of FPGA resources utilization}
\label{tbl_network_fpga}
\resizebox{\columnwidth}{!}{
\begin{tabular}{c|c|c}
\hline
\bf{Type} & \bf{RLF-based Network} & \bf{BNNWallace-based Network}\\\hline
Total ALMs & 98,006/113,560 ($86.3\%$) &  91,126/113,560 ($80.2\%$)\\
Total DSPs & 342/342 ($100\%$)& 342/342 ($100\%$) \\
Total Registers & 88,720 & 78,800 \\
Total Block Memory Bits & 4,572,928/12,492,800 ($36.6\%$)& 4,880,128/12,492,800 ($39.1\%$)  \\
\hline
	\end{tabular}
}
\end{table}

\begin{table}[t]
\centering
\caption{Performance comparison on the MNIST dataset}
\label{tbl:perfcon}
\resizebox{\columnwidth}{!}{
\begin{tabular}{c|c|c}
\hline
\bf{Configuration} & \bf{Throughtput (Images/s)} & \bf{Energy (Images/J) }\\\hline
Intel i7-6700k & 10478.1 &  115.1 \\
Nvidia GTX1070 & 27988.1 & 186.6 \\
RLF-based FPGA Implementation & 321543.4 & 52694.8 \\
BNNWallace-based FPGA Implementation & 321543.4 & 37722.1 \\
\hline
	\end{tabular}
}
\end{table}

\begin{table}
	\centering
	\caption{Accuracy Comparison on the MNIST Dataset}
	\label{tbl:acc}
	\begin{tabular}{c|c}
		\hline
\bf{Model} & \bf{Testing Accuracy}\\\hline
FNN+Dropout (Software)  & $97.50\%$ \\
BNN (Software)          & $98.10\%$ \\
VIBNN (Hardware)        & $97.81\%$ \\
\hline
	\end{tabular}
\end{table}

\begin{table}[t]
\centering
\caption{Accuracy comparison on classification tasks}
\label{tbl:med}
\resizebox{\columnwidth}{!}{
\begin{tabular}{c|c|c|c}
\hline
\bf{Dataset} & \bf{FNN (Software)} & \bf{BNN (Software)} & \bf{VIBNN (Hardware)}\\\hline
Parkinson Speech Dataset (Modified)    & $60.28\%$  &  $95.68\%$  &  $95.33\%$\\
Parkinson Speech Dataset (Original)    & $85.71\%$) &  $95.23\%$  &  $94.67\%$\\
Diabetics Retinopathy Debrecen Dataset & $70.56\%$  &  $75.76\%$  &  $75.21\%$\\
Thoracic Surgery Dataset               & $76.69\%$  &  $82.98\%$  &  $82.54\%$\\
TOX21:NR.AhR                           & $91.10\%$  &  $90.42\%$  &  $90.11\%$\\
TOX21:SR.ARE                           & $83.41\%$  &  $83.24\%$  &  $83.01\%$\\
TOX21:SR.ATAD5                         & $93.36\%$  &  $94.05\%$  &  $93.67\%$\\
TOX21:SR.MMP                           & $89.69\%$  &  $88.76\%$  &  $88.43\%$\\
TOX21:SR.P53                           & $91.88\%$  &  $93.33\%$  &  $92.87\%$\\
\hline
	\end{tabular}
}
\end{table}

\subsection{Accuracy: Classification Tasks}
In this work, we test the classification performance of the BNNs with several classification tasks on both software and hardware platforms, comparing with the performance of FNNs. The first task is the image classification on MNIST dataset. As discussed in the previous subsection, the network we use has two hidden layers of neurons apart from the 784 ($28\times28$) inputs and 10 outputs. Compared with the 8-bit fixed point representation used in hardware, the BNN implementation in software uses 32-bit floating point for all operations. Results of accuracy testing are reported in Table \ref{tbl:acc}. The BNN implemented in software has higher classification accuracy than that of FNNs with dropout applied. The proposed implementation of BNN on FPGA degrades only 0.29\% accuracy compared to its software model. 

To further evaluate accuracy performance, we test on several disease diagnosis datasets including the Parkinson Speech Dataset\cite{6451090}, the Diabetics Retinopathy Debrecen Dataset\cite{Antal:2014:ESA:2611848.2612138}, Thoracic Surgery Dataset\cite{zieba2013boosted} and part of the TOX21 dataset\cite{attene2013tox21} in which multiple features of chemicals are given to detect toxic compound. For the parkinson speech dataset, we relocate part of the training data for testing to create a small data traning scenario(shown as Parkinson Speech Dataset(modified) in Table \ref{tbl:med}). As Shown in Table \ref{tbl:med}, BNNs can achieve similar or better accuracy  compared to FNNs, especially for small datasets. The accuracy of VIBNN degrades very little compared to the software implementation of BNN.

\section{Conclusion}
In this paper, we propose VIBNN, 
an FPGA-based hardware accelerator design for variational inference on BNNs. 
We explore the design space for massive amount of  Gaussian variable sampling tasks in BNNs.
Specifically, we introduce two high performance parallel Gaussian random number generators: 
{\em 1)} the RAM-based Linear Feedback Gaussian Random Number Generator (RLF-GRNG), 
which is inspired by the properties of binomial distribution and linear feedback logics; and 
{\em 2)} the Bayesian Neural Network-oriented Wallace Gaussian Random Number Generator. 
To achieve high scalability and efficient memory access,
we propose a deep pipelined accelerator architecture 
with fast execution and good hardware utilization.  
The proposed VIBNN achieves high throughput of 321,543.4 Images/s and high energy efficiency upto 52,694.8 Images/J. Experimental results suggest that the proposed VIBNN can achieve similar accuracy performance as software implemented BNNs.

\begin{acks}
This paper is in part supported by National Science Foundation CNS-1704662, CCF-1717754, CCF-1750656, and Defense Advanced Research Projects Agency (DARPA) MTO seedling project.
\end{acks}

\bibliographystyle{ACM-Reference-Format}

\end{document}